\documentclass[10pt,twocolumn,letterpaper]{article}

\usepackage[pagenumbers]{cvpr} 

\definecolor{cvprblue}{rgb}{0.21,0.49,0.74}

\usepackage[ruled, linesnumbered]{algorithm2e}
\SetAlCapHSkip{0pt} 
\usepackage{etoolbox}
\makeatletter
\patchcmd{\algocf@makecaption@ruled}{\hsize}{\textwidth}{}{} 
\patchcmd{\@algocf@start}{-1.5em}{0em}{}{} 
\makeatother

\usepackage[pagebackref,breaklinks,colorlinks,allcolors=cvprblue]{hyperref}
\usepackage{wrapfig}
\usepackage{graphicx}
\usepackage{cuted}
\usepackage{multirow}
\usepackage{capt-of}
\usepackage{makecell}
\usepackage{natbib}
\usepackage{amsmath}
\usepackage{amsfonts}
\usepackage{bbold}
\usepackage{array}
\usepackage{hyperref}
\usepackage{titling}

\def\paperID{10124} 
\def\confName{CVPR}
\def\confYear{2025}

\title{\textbf{EditSplat: Multi-View Fusion and Attention-Guided Optimization for View-Consistent 3D Scene Editing with 3D Gaussian Splatting}}
\date{}

\author{
Dong In Lee\textsuperscript{\normalfont 1} \quad
Hyeongcheol Park\textsuperscript{\normalfont 1} \quad
Jiyoung Seo\textsuperscript{\normalfont 1}\quad
Eunbyung Park\textsuperscript{\normalfont 2} \quad
Hyunje Park\textsuperscript{\normalfont 1}\\
Ha Dam Baek\textsuperscript{\normalfont 1}\quad
Sangheon Shin\textsuperscript{\normalfont 3}\quad
Sangmin Kim\textsuperscript{\normalfont 3}\quad
Sangpil Kim\textsuperscript{\normalfont 1}\thanks{Corresponding author.}\vspace{0.6em}\\
\textsuperscript{\normalfont 1}Korea University \quad
\textsuperscript{\normalfont 2}Yonsei University \quad
\textsuperscript{\normalfont 3}Hanhwa Systems
\vspace{-1.em}
}

\begin{document}
\maketitle


\begin{strip}\centering
\vspace{-4.5em}
\includegraphics[width=\textwidth]{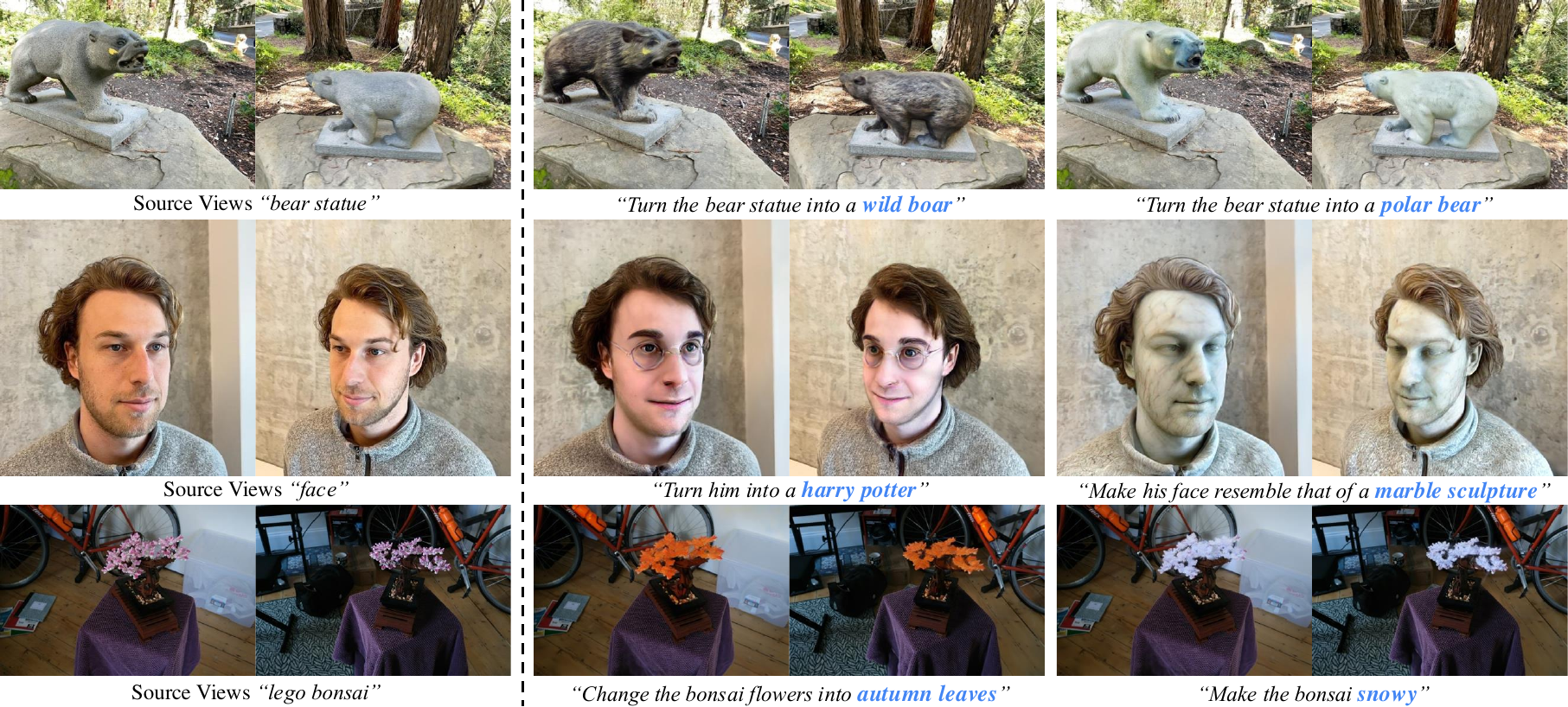}
\vspace{-2.em}
\captionof{figure}
{\textbf{Result of EditSplat}. EditSplat enables flexible and high-quality editing of pre-trained 3D Gaussian Splatting models guided solely by textual instructions. Through its design focused on multi-view consistency, efficient optimization, and precise semantic local editing, Our approach demonstrates robust performance, producing realistic and fine-grained 3D scene modifications.
}

\label{fig:teaser}
\end{strip}


\begin{abstract}

Recent advancements in 3D editing have highlighted the potential of text-driven methods in real-time, user-friendly AR/VR applications. However, current methods rely on 2D diffusion models without adequately considering multi-view information, resulting in multi-view inconsistency. While 3D Gaussian Splatting (3DGS) significantly improves rendering quality and speed, its 3D editing process encounters difficulties with inefficient optimization, as pre-trained Gaussians retain excessive source information, hindering optimization. To address these limitations, we propose \textbf{EditSplat}, a novel text-driven 3D scene editing framework that integrates Multi-view Fusion Guidance (MFG) and Attention-Guided Trimming (AGT). Our MFG ensures multi-view consistency by incorporating essential multi-view information into the diffusion process, leveraging classifier-free guidance from the text-to-image diffusion model and the geometric structure inherent to 3DGS. Additionally, our AGT utilizes the explicit representation of 3DGS to selectively prune and optimize 3D Gaussians, enhancing optimization efficiency and enabling precise, semantically rich local editing. Through extensive qualitative and quantitative evaluations, EditSplat achieves state-of-the-art performance, establishing a new benchmark for text-driven 3D scene editing.
Project website: \url{https://kuai-lab.github.io/editsplat2024/}
\vspace{-1.5em}
\end{abstract}    
\section{Introduction}

Text-driven 3D scene editing, which enables manipulation of 3D scenes using only text instructions, is gaining research momentum. The goal is to edit 3D representations accurately and efficiently through text prompts, facilitating real-time, user-friendly 3D editing for film and game development, digital content creation, or AR/VR applications. 

\begin{figure}[t]
    \centering
    \includegraphics[width=\linewidth]{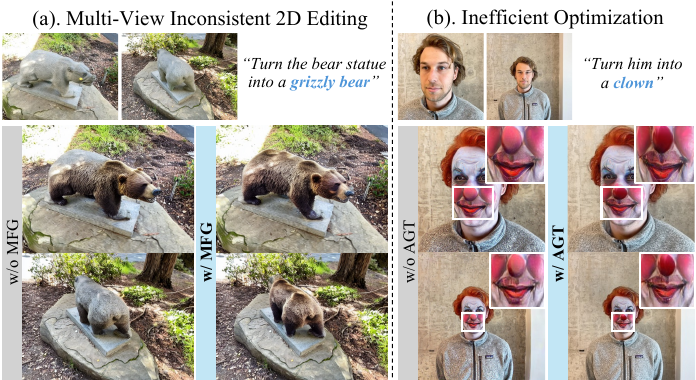}
    \vspace{-2.em}
    \caption{
    Challenging cases. (a) Conventional 2D editing without Multi-view Fusion Guidance (MFG) results in inconsistent textures across different views (e.g., bear fur), whereas applying MFG achieves consistent multi-view edits. (b) Editing without Attention-Guided Trimming (AGT) leads to inefficient optimization, resulting in less edited regions (e.g., clown’s nose). AGT effectively enhances optimization quality, producing richer colors.
    }
    \label{fig:intro}
    \vspace{-1.7em}
\end{figure}

InstructNeRF2NeRF (IN2N)~\cite{haque2023instruct} introduced the pipeline for text-driven 3D editing by leveraging 2D diffusion model~\cite{rombach2022high}, such as InstructPix2Pix~\cite{brooks2023instructpix2pix}. Subsequent methods~\cite{wu2024gaussctrl, chen2024gaussianeditor, wang2025view, mirzaei2025watch}, inspired by IN2N, iteratively edit rendered images while updating the underlying 3D representation. However, these approaches independently edit single views with 2D diffusion models, neglecting multi-view consistency, as illustrated in \cref{fig:intro}-(a). More recent methods~\cite{dong2024vica, chen2024dge, Chen_2024_CVPR} incorporate multi-view information, but they lack a comprehensive strategy for consistent view alignment. For example, \cite{dong2024vica, chen2024dge} use only a limited set of key views, requiring additional diffusion passes or extra-parameterized layers to integrate multi-view information. Similarly, \cite{Chen_2024_CVPR} trains a diffusion model to achieve multi-view consistency, incurring high costs. 
Consequently, existing methods still exhibit multi-view inconsistency, resulting in noisy gradients that hinder optimization and produce suboptimal outputs, such as minimal edits or blur outputs.


3D Gaussian Splatting (3DGS)~\cite{kerbl20233d} has emerged as a foundational model in 3D representation, surpassing Neural Radiance Field (NeRF)~\cite{mildenhall2021nerf} in rendering quality and speed. Unlike NeRF's implicit representations, 3DGS employs explicit anisotropic ellipsoids, enabling faster training and high-quality reconstruction. However, editing a pre-trained 3DGS model introduces inefficiency, as pre-trained Gaussians tend to retain source visual and geometric details excessively, impeding efficient convergence as illustrated in \cref{fig:intro}-(b). This underscores the need to manage redundant Gaussians for more effective optimization.

To tackle the limitations in both the multi-view inconsistency and the optimization inefficiency of pre-trained Gaussians in editing, we propose a novel 3D scene editing framework, \textbf{EditSplat}. To ensure multi-view consistency between edited images, we propose a \textit{Multi-view Fusion Guidance (MFG)} method that ensures multi-view consistent editing utilizing a 2D diffusion model and the geometric structure inherent to 3DGS. Inspired by \cite{dong2024vica, Chen_2024_CVPR}, our method projects initially edited multi-view images onto a target view using 3DGS depth maps, enabling smooth blending based on depth values across views to integrate multi-view information. Leveraging classifier-free guidance in the text-image diffusion process~\cite{ho2022classifier, brooks2023instructpix2pix}, we carefully edit source images based on text prompt by incorporating multi-view fused images into the diffusion process, ensuring robust alignment of edited images with multi-view details. To further preserve source fidelity, we utilize the source images as auxiliary guidance.    By balancing guidance scores between multi-view fused images, source images, and text prompts, our method enables the diffusion model to edit source images to the multi-view-consistent outputs, aligning with essential multi-view information. Unlike iterative editing approaches that require expensive optimization processes during rendered image editing, our proposed EditSplat edits source images directly, allowing for immediate 3DGS model optimization.

Additionally, we propose an \textit{Attention-Guided Trimming (AGT)} method to improve the optimization efficiency of pre-trained Gaussians and enable semantically rich local editing. Utilizing the explicit representation of 3DGS, we assign attention weight to each Gaussian based on the attention map from the diffusion model. High attention weights highlight semantically meaningful regions, indicating substantial visual and geometric changes required for effective edits. Thus, pre-trained Gaussians with high attention weight generally need significant changes with the optimization process and become redundant during the densification process as the number of Gaussians retaining excessive source information increases. To address this problem, we prune a suitable proportion of pre-trained Gaussians with high attention weight before editing, allowing the remaining Gaussians to achieve efficient densification and optimization. In addition, we selectively optimize Gaussians with high attention weight, enabling semantic local editing. Overall, the proposed AGT simplifies optimization and improves semantic precision, enabling faster and more efficient 3DGS editing.

\begin{figure*}[t]
    \vspace{-2.2em}
    \centering
    \includegraphics[width=\linewidth]{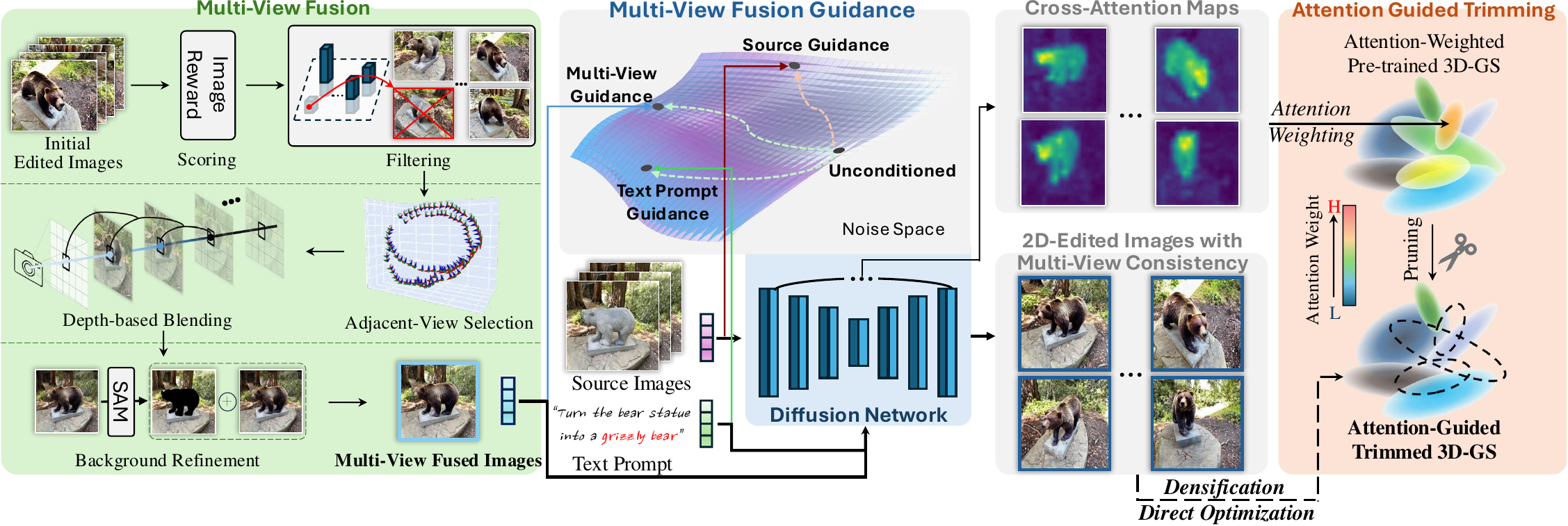}
    \vspace{-2.em}
    \caption{
    \textbf{EditSplat Overview.} EditSplat consists of two main methods: (1) Multi-view Fusion Guidance (MFG, \cref{sec:MFG}), which aligns multi-view information with text prompts and source images to ensure multi-view consistency; (2) Attention-Guided Trimming (AGT, \cref{sec:AGT}), which prunes pre-trained Gaussians for optimization efficiency and selectively optimizes Gaussians for semantic local editing.
    }
    \label{fig:overview}
    \vspace{-1.em}
\end{figure*}

We conduct extensive qualitative and quantitative experiments to validate the high quality and effectiveness of EditSplat. Our key contributions are as follows:

\begin{itemize}

\item We propose a Multi-view Fusion Guidance (MFG) method that integrates multi-view information into the diffusion process to achieve robust alignment with multi-view details, ensuring consistent multi-view editing.

\item We introduce an Attention-Guided Trimming (AGT) technique that prunes 3D Gaussians with high attention weight and selectively optimizes them, enhancing optimization efficiency and enabling semantic local editing.

\item Through extensive qualitative and quantitative evaluations, we demonstrate that EditSplat outperforms existing methods in both performance and effectiveness.

\end{itemize}

\section{Related Work}

\paragraph{2D Image Editing.}

Despite advancements in 3D generative models~\cite{liu2024direct, deitke2023objaverse, chang2015shapenet, Shue_2023_CVPR, anciukevivcius2023renderdiffusion, xu2024dmvd, lin2023magic3d, karnewar2023holodiffusion}, their application in 3D editing is constrained by limited 3D data availability. To overcome this limitation, recent studies have shifted towards leveraging large-scale 2D generative models~\cite{ho2020denoising, song2021denoising, peebles2023scalable, ho2022cascaded, saharia2022palette, ramesh2022hierarchical}, such as Text-to-Image (T2I) diffusion models~\cite{rombach2022high, zhang2023adding, xu2023inversion, si2024freeu}, to enable more flexible 3D editing. DreamBooth~\cite{ruiz2023dreambooth} proposes subject-specific editing by fine-tuning the T2I diffusion model Imagen~\cite{saharia2022photorealistic}. Delta Denoising Score (DDS)~\cite{hertz2023delta} utilizes Score Distillation Sampling (SDS)~\cite{poole2022dreamfusion} for precise image modifications. Prompt-to-Prompt (P2P)~\cite{hertz2022prompt} harnesses cross-attention maps in the diffusion process for detailed edits. InstructPix2Pix (IP2P)~\cite{brooks2023instructpix2pix} enables instruction-driven editing by fine-tuning Stable Diffusion~\cite{rombach2022high} on a synthetic dataset of image-text pairs generated using GPT-3~\cite{brown2020language} and P2P~\cite{hertz2022prompt}. Our EditSplat utilizes IP2P as a 2D image editor. Utilizing its image guidance capability through classifier-free guidance, EditSplat effectively integrates multi-view information into the diffusion process, facilitating multi-view consistent editing for enhancing 3D editing.

\vspace{-1.2em}
\paragraph{Text-driven 3D Scene Editing.}
Text-driven 3D scene editing aims to modify 3D representations precisely and efficiently using textual instructions. Despite recent progress, key challenges such as multi-view consistency, semantic localization, and optimization efficiency remain.

Recent advancements~\cite{park2023ed, zhuang2023dreameditor, li2024focaldreamer, koo2024posterior} employ Score Distillation Sampling (SDS)~\cite{poole2022dreamfusion}, while iterative methods such as InstructNerf2Nerf~\cite{haque2023instruct} and its extensions~\cite{chen2024gaussianeditor, wu2024gaussctrl, wang2025view, mirzaei2025watch, luo2024trame} repeatedly edit rendered images while optimizing the underlying 3D representation. However, those works often overlook multi-view information, leading to inconsistency.

Recent methods explicitly addressing multi-view consistency, VICA-NeRF~\cite{dong2024vica} and ConsistDreamer~\cite{Chen_2024_CVPR}, employ view projection strategies but suffer from noisy projection outputs and computational overhead due to additional diffusion steps~\cite{dong2024vica} or extensive U-Net~\cite{ronneberger2015u} training~\cite{Chen_2024_CVPR}. Other approaches~\cite{chen2024dge, karim2023free} utilize the epipolar lines, though they still struggle to preserve fine-grained details. Moreover, 3DEgo~\cite{khalid20253dego} employs classifier-free guidance with IP2P~\cite{brooks2023instructpix2pix}, but its requirement for guidance from all frames in single-view edits inevitably results in inconsistencies due to misalignment viewpoint.

Furthermore, most existing methods~\cite{chen2024gaussianeditor, chen2024dge, wu2024gaussctrl, dong2024vica, khalid20253dego} rely on binary masks from the source image to localize the editing area, which constrains edits and overlooks semantic information when the target object is not clearly defined.

For multi-view consistency, we introduce a Multi-view Fusion Guidance (MFG) involving sequential projection and blending, avoiding noise issues and the additional training~\cite{Chen_2024_CVPR} or diffusion refinement~\cite{dong2024vica}. Regarding semantic localization, inspired by DreamEditor~\cite{zhuang2023dreameditor}, EditSplat explicitly assigns attention maps to 3D Gaussians, achieving higher semantic precision without binary mask constraints. Additionally, as far as we know, we are the first to address the optimization inefficiency inherent in pre-trained 3DGS models by strategically pruning 3D Gaussians, enhancing both densification and optimization efficiency. 
\begin{figure*}[t]
    \vspace{-1.em}
    \centering
    \includegraphics[width=\linewidth]{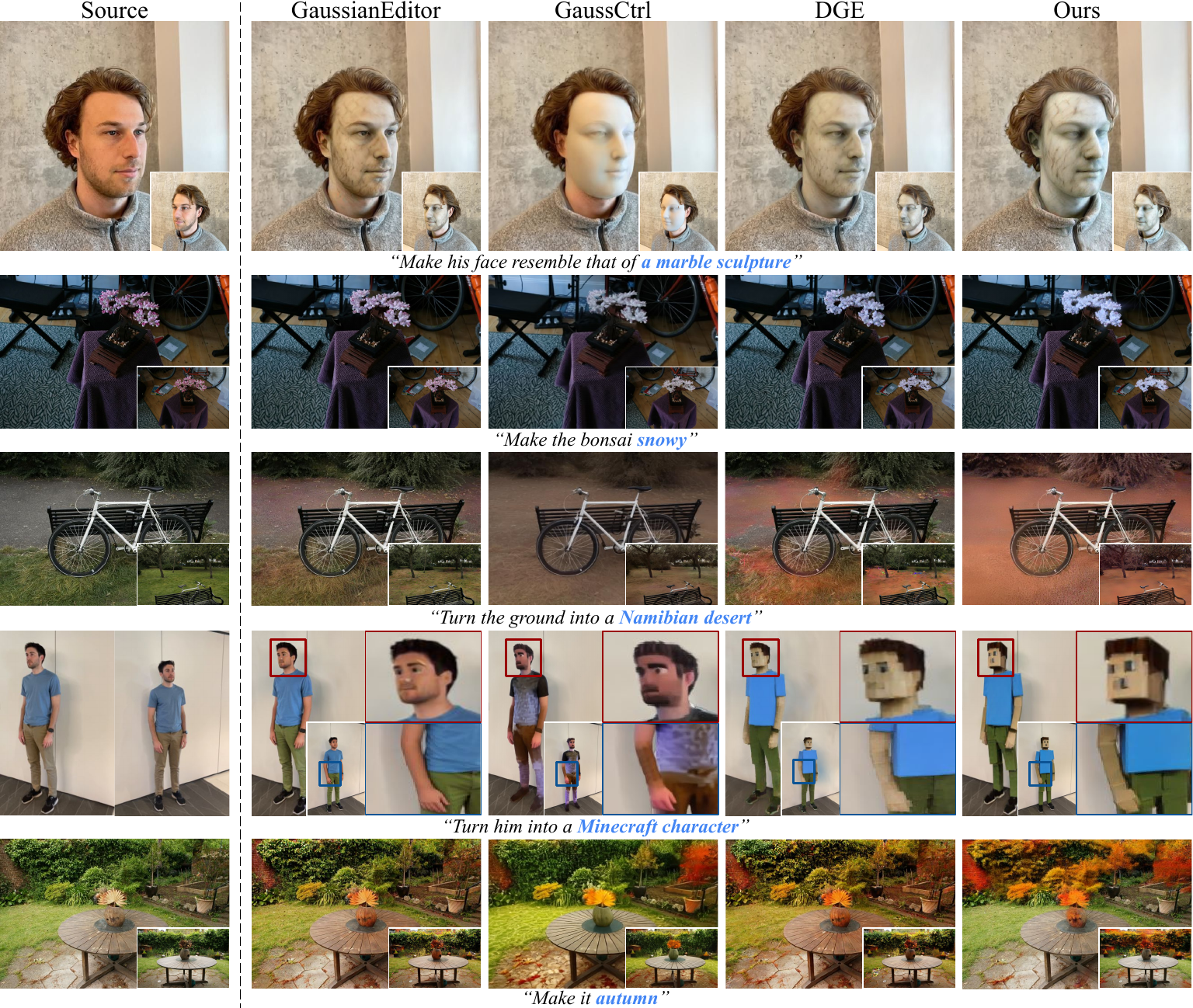}
    \vspace{-2.em}
    \caption{
    \textbf{Qualitative Comparison}. EditSplat provides more intense and precise editing compared to other baselines. The leftmost column shows source images, while the right columns show rendering images from edited 3DGS. In each corner of the images, we include different views of the corresponding image to compare multi-view consistency. Note that our EditSplat outperforms both local and global editing.
    }
    \label{fig:comparison}
    \vspace{-1.5em}
\end{figure*}

\section{Method}

We propose a novel 3D editing framework, EditSplat, carefully designed to achieve three key objectives: (i) multi-view consistent editing, (ii) efficient optimization, and (iii) semantically rich local editing. To meet these goals, we introduce Multi-view Fusion Guidance (MFG) and Attention-Guided Trimming (AGT). MFG incorporates essential multi-view information into the diffusion model to ensure multi-view consistency. In AGT, attention weights from the diffusion model are explicitly assigned to each Gaussian in 3D Gaussian Splatting (3DGS). This enables efficient convergence by pruning Gaussians with high attention weight and selectively optimizing them for semantic local editing. The overall pipeline is shown in \cref{fig:overview}. In the following sections, we provide a review of 3DGS and classifier-free guidance in \cref{sec:3-1}, followed by our primary methods: MFG in \cref{sec:MFG} and AGT in \cref{sec:AGT}.


\vspace{-0.1em}
\subsection{Preliminaries} 
\label{sec:3-1}

\vspace{-0.2em}
\paragraph{3D Gaussian Splatting.} 

3D Gaussian Splatting~\cite{kerbl3Dgaussians} is an explicit 3D scene representation using anisotropic Gaussians to model complex structures. A set of 3D Gaussians \( G \) represents the scene, where each 3D Gaussian \( g \in G \) is defined by a center \( \mu \in \mathbb{R}^3 \), a covariance matrix \( \Sigma \in \mathbb{R}^{3 \times 3} \), spherical harmonic coefficients \( f \in \mathbb{R}^k \) (with \( k \) indicating the degrees of freedom), and opacity \( \sigma \in \mathbb{R} \). The covariance matrix \( \Sigma \) is composed of a rotation matrix \( R \) and a scaling matrix \( S \) as \( \Sigma = R S S^T R^T \). For rendering, a 3D Gaussian function is defined as

\vspace{-0.8em}
\begin{equation}
    g(x ; \mu, \Sigma) = e^{-\frac{1}{2} (x-\mu)^T \Sigma^{-1} (x-\mu)}.
\end{equation}
Each 3D Gaussian \( g \) is projected into a 2D image space using a world-to-image projective transform, with its Jacobian evaluated at \( \mu \).
The final color of each pixel is calculated by blending contributions from Gaussians along a ray:
\vspace{-0.8em}
\begin{equation}
    C = \sum_{i \in N} c_i \alpha_i \prod_{j=1}^{i-1} (1 - \alpha_j),
\vspace{-0.6em}
\end{equation}
where blending weight \( \alpha_i \in \mathbb{R} \) is defined as the density value of \(g\) multiplied by \(\sigma \), and \(c \in \mathbb{R}^3 \) is color of the Gaussian. This tile-based approach supports real-time rendering, with Gaussian parameters \( \mu \), \( \Sigma \), \( f \), and \( \sigma \) optimized to represent scenes accurately through photometric loss.

\vspace{-1.em}
\paragraph{Classifier-free Guidance.}
Classifier-free diffusion guidance~\cite{ho2022classifier} improves the alignment between generated outputs and conditioning prompts in tasks, such as image- or text-conditional image generation by guiding sampling toward regions where the implicit classifier \( p_{\theta}(h|z_t) \) assigns high probability to the target condition \( h \). In models with multiple guidance inputs, such as InstructPix2Pix~\cite{brooks2023instructpix2pix} (IP2P) used in our approach, the score network \( \epsilon_{\theta}(z_t, h_I, h_T) \) incorporates both an input image \( h_I \) and a text instruction \( h_T \) for guide generation. Here, \( z_t \) denotes the latent representation of the input image at timestep \( t \) with added noise, and \( {\epsilon_\theta} \) represents the noise prediction network (e.g., U-net~\cite{ronneberger2015u}).


To control the influence of each conditioning, guidance scales \( s_I \) and \( s_T \) are introduced, adjusting the extent to which the generated outputs align with the input image and text instruction. The guided prediction \( \tilde{\epsilon}_{\theta} \) is defined as:
\begin{small}
\begin{align}
    \tilde{\epsilon}_{\theta}(z_t, h_I, h_T) &= \epsilon_{\theta}(z_t, \varnothing, \varnothing) \notag \\ &\quad + s_I  (\epsilon_{\theta}(z_t, h_I, \varnothing) - \epsilon_{\theta}(z_t, \varnothing, \varnothing)) \notag \\
    &\quad + s_T  (\epsilon_{\theta}(z_t, h_I, h_T) - \epsilon_{\theta}(z_t, h_I, \varnothing)),
\label{eq:CFG}
\end{align}
\end{small}
where \( s_I \) and \( s_T \) modulate the influence of the image condition \( h_I \) and text instruction \( h_T \). This balances conditioned and unconditioned score estimates to guide generation, prioritizing one or both types of conditioning.




\subsection{Multi-View Fusion Guidance (MFG)}
\label{sec:MFG}

Recent studies~\cite{chen2024gaussianeditor, haque2023instruct, wu2024gaussctrl, dong2024vica, chen2024dge} struggle with consistent editing across views due to limited consideration of multi-view information, resulting in multi-view inconsistency that causes minimal edits or blurriness. To overcome this challenge, we focus on integrating essential multi-view details into the 2D diffusion model, enabling consistent editing.


\vspace{-1.em}
\paragraph{Multi-View Fusion.} \hspace{-0.4em}
We first edit the whole source images using the diffusion model. 
Inspired by ~\cite{dong2024vica, Chen_2024_CVPR}, we incorporate multi-view information by projecting and blending the initially edited multi-view images onto each target view using depth maps from 3DGS. To enhance output quality, we rank these images using ImageReward~\cite{xu2024imagereward}, which scores fidelity and text alignment based on human feedback. Since some initial edits are generally misaligned, we exclude the bottom 15\% of low-scoring images from projection. 

For the multi-view projection process, we select the top 5 adjacent views, prioritizing them based on their proximity to the target view, determined by camera position and orientation in world coordinates. To manage overlapping pixels across multiple views, we implement an iterative alpha blending strategy guided by depth values. Starting from the farthest pixel (i.e., highest depth value), pixel pairs are blended iteratively in decreasing depth order. The final pixel color is determined by aligning and unprojecting the multi-view images into 3D space using depth maps and camera parameters, followed by reprojecting onto the target view to resolve overlapping pixel contributions. This approach ensures that each target view captures comprehensive details shared across multiple views.
To handle sparse background regions and achieve seamless integration, we refine these areas using SAM~\cite{kirillov2023segment} to fill them with source content. This sequential multi-view fusion strategy enriches each target view with consistent and detailed information, reinforcing alignment and coherence across all views. The resulting multi-view fusion image \(h_M\) for each target view serves as an essential input for subsequent editing processes, integrating comprehensive visual and geometric details shared across the views. Note that further details are provided in the supplementary material.

\vspace{-1.2em}
\paragraph{Alignment with Multi-View Information.}
To effectively integrate multi-view fusion information into the diffusion process and edit source images without additional training~\cite{Chen_2024_CVPR}, extra-parameterized layers~\cite{chen2024dge}, or repeated diffusion passes~\cite{dong2024vica}, we leverage classifier-free guidance. This allows the direct incorporation of multi-view fusion details into the diffusion process. According to~\citet{liu2022compositional}, a conditional diffusion model can combine score estimates from multiple conditioning values. Building on this, we adapt this approach, facilitating the implicit classifier \( p_\theta \) to assign high probabilities to multi-view fused visual features, color consistency, and structural properties. This ensures the edited images maintain a strong alignment with multi-view fused images throughout the editing process. 

Specifically, we provide the multi-view fusion image \( h_M \) as a guidance conditioning during source image editing. This enhances alignment with multi-view details, ensuring multi-view consistent editing. To maintain source fidelity, we include the original source image \( h_S \) as auxiliary guidance, preserving its content. Our guided score prediction \( \tilde{\epsilon_\theta} \) is defined as follows:

\vspace{-1.em}
\begin{small}
\begin{equation}
\begin{aligned}
    \tilde{\epsilon_\theta}\left(z_t, h_S, h_T, h_M\right) & =  \; {\epsilon_\theta}\left(z_t, \varnothing, \varnothing\right) \\
    & + s_T   \left({\epsilon_\theta}\left(z_t, h_M, h_T\right) - {\epsilon_\theta}\left(z_t, h_M, \varnothing\right)\right) \\
    & + s_M   \left({\epsilon_\theta}\left(z_t, h_M, \varnothing\right) - {\epsilon_\theta}\left(z_t, \varnothing, \varnothing\right)\right) \\
    & + s_S  \left({\epsilon_\theta}\left(z_t, h_S, \varnothing\right) - {\epsilon_\theta}\left(z_t, \varnothing, \varnothing\right)\right),
\end{aligned}
\label{eq:MFG}
\end{equation}
\end{small}
where \( h_M \), \( h_S \), and \( h_T \) correspond to the multi-view fusion image, source image, and text prompt, respectively. Each guidance input is modulated by its respective scale factor: \( s_M \) for Multi-view Fusion Guidance, \( s_S \) for the source image, and \( s_T \) for the text prompt.

Our method efficiently directs the diffusion process toward generating outputs that achieve seamless multi-view consistency, including multi-view fusion images' texture, color, geometric, and source images. By balancing guidance scores between \( s_M \), source images \( s_S \), and text prompts \( s_T \), our approach enables the diffusion model to align edits with essential multi-view information and generate consistent outputs.
Unlike iterative editing approaches~\cite{chen2024gaussianeditor, haque2023instruct, wang2025view, wu2024gaussctrl} that iteratively edit rendered images while updating the underlying 3D representation, our approach directly edits the source image, ensuring precise alignment with multi-view information and the text prompt.

\subsection{Attention-Guided Trimming (AGT)}
\label{sec:AGT}
When editing a 3DGS, optimization inefficiency occurs as pre-trained Gaussians tend to excessively retain source attributes, obstructing effective optimization. Additionally, previous methods~\cite{chen2024gaussianeditor, chen2024dge, dong2024vica, wu2024gaussctrl} using 2D binary masks from SAM~\cite{kirillov2023segment} for local editing often overlook semantic regions, as these masks are extracted solely from the source image rather than the edited output. This local editing issue can lead to suboptimal results when the target object is not clearly defined. These underlines the need to manage Gaussians in editing. To improve efficiency and semantic local editing, we trim Gaussians by pruning redundant Gaussians and selectively optimizing those relevant, guided by attention maps from the diffusion model. 

\vspace{-1.em}
\paragraph{Attention Map.}
Attention maps from the text-to-image diffusion U-Net’s cross-attention layers highlight regions that require intensified focus for generation~\cite{hertz2022prompt,liu2024towards}, revealing the correlation between individual words and generated image regions, computed as \(\text{Softmax}\left(\frac{QK^T}{\sqrt{q}}\right)\), where \( Q \) is the query matrix projected from the spatial features of the noisy image with dimension \( q \), and \( K \) is the key matrix projected from the textual embedding. In image editing diffusion (IP2P), high attention weights similarly indicate areas where edits should be concentrated, typically aligning with regions of significant visual change.

\vspace{-1.em}
\paragraph{Attention Weighting 3D Gaussians.}

To assign attention weights to each pre-trained Gaussian, we first unproject multi-view-consistent cross-attention maps onto the 3D Gaussians through inverse rendering~\cite{chen2024gaussianeditor}. These maps, resized to rendering size using bilinear interpolation, correspond to specific word tokens for the editing target in the MFG editing process. Given the \( j \)-th Gaussian in the 3DGS, we compute its cumulative attention weight \( w_j \) by aggregating the attention map from each view \( v \) of all views \( V \) as:

\vspace{-1.2em}
\begin{equation}
     w_j = \frac{1}{ \sum_{v \in V} |S_{v,j}|} \sum_{v \in V} \sum_{s \in S_{v,j}} \text{Softmax}\left(\frac{Q_vK^T}{\sqrt{q}}\right)_{s}, 
\label{eq:attnweighting}
\end{equation}
where \(S_{v,j}\) denotes a set of indices of all attention weights affected by the \(j\)-th Gaussian from the attention map for the $v$-th view.
$\text{Softmax}\left( \frac{Q_vK^T}{\sqrt{q}} \right)_s \in [0,1]$ denotes an individual attention weight at $s$ from an attention map for the $v$-th view.
By aggregating attention weight across \( V \), this weighting mechanism ensures that each Gaussian reflects the overall importance of multi-view attention maps, emphasizing regions that require pronounced geometrical or visual changes during editing. Details on preparing the attention map are in the supplementary material.


\vspace{-1.em}
\paragraph{Trimming 3D Gaussians.}

After weighting attention map to each Gaussian, we apply thresholding to obtain \( w_j' = w_j \mathbb{1}\{ w_j \geq w_\text{thres} \} \) to exclude Gaussians with low attention from the optimization by setting those gradients to zero, directing updates solely toward Gaussians with high attention. This selective optimization enhances precision by concentrating on semantically rich regions and aligning edits with meaningful word-image correlations.


Pre-trained Gaussians with high attention weights hinder optimization for editing, as these regions indicate areas requiring substantial visual and geometric modifications. This inefficiency is intensified as the number of Gaussians retaining excessive source information increases during the densification process.

To improve densification and optimization efficiency, we prune an optimal proportion of Gaussians with high attention before training, focusing on essential areas for more effective editing.



\vspace{-1.3em}
\begin{equation}
    G_{\text{pruning}} = \{ g_j \mid g_j \in G, w_j' \geq \text{Top-}k\% (G) \},
\label{eq:attnpruning}
\end{equation}
where $\text{Top-}k\%(G)$ returns the threshold weight for top \( k\% \) Gaussian and then \(G_{\text{pruning}} \) consists of Gaussians selected for pruning, the top \( k\% \) of Gaussians in \(G \) based on attention weight \( w_j' \).
By removing \( G_{\text{pruning}} \), this pruning step eliminates redundant pre-trained Gaussians with high attention weight, allowing efficient densification and optimization to concentrate on the remaining necessary Gaussians and thereby enhancing overall efficiency.

\vspace{-1.1em}
\paragraph{Optimization for 3D Editing.}
Given the source Gaussians trimmed by AGT, \(G_\text{agt} = G \setminus G_\text{pruning} \), and its rendering function \( \hat{I}_v \), as well as the multi-view consistent edited image \( I^{mfg}_v \) by MFG editing process, the final optimized Gaussians \( G_{\text{edit}} \) is obtained by minimizing the editing loss across all views \( V \):

\vspace{-0.5em}
\begin{equation}
    G_{\text{edit}} = \arg\min_{G_\text{agt}} \sum_{v \in V} \mathcal{L}_{\text{edit}}(\hat{I}_v(G_\text{agt}), I_v^\text{mfg}),
\label{eq:loss}
\end{equation}
where \( \mathcal{L}_{\text{edit}} \) measures the discrepancy between rendered image \( \hat{I}_v(G_\text{agt}) \) and \( \hat{I}^{mfg}_v\) for each view \( v \). We use combination of LPIPS~\cite{Zhang_2018_CVPR} and \( L_1 \) loss, weighted at a 1:1 ratio, as commonly empolyed in 3D scene editing task~\cite{haque2023instruct}.

\section{Experiments}


\begin{figure*}[t]
    \vspace{-1.5em}
    \centering
    \includegraphics[width=\linewidth]{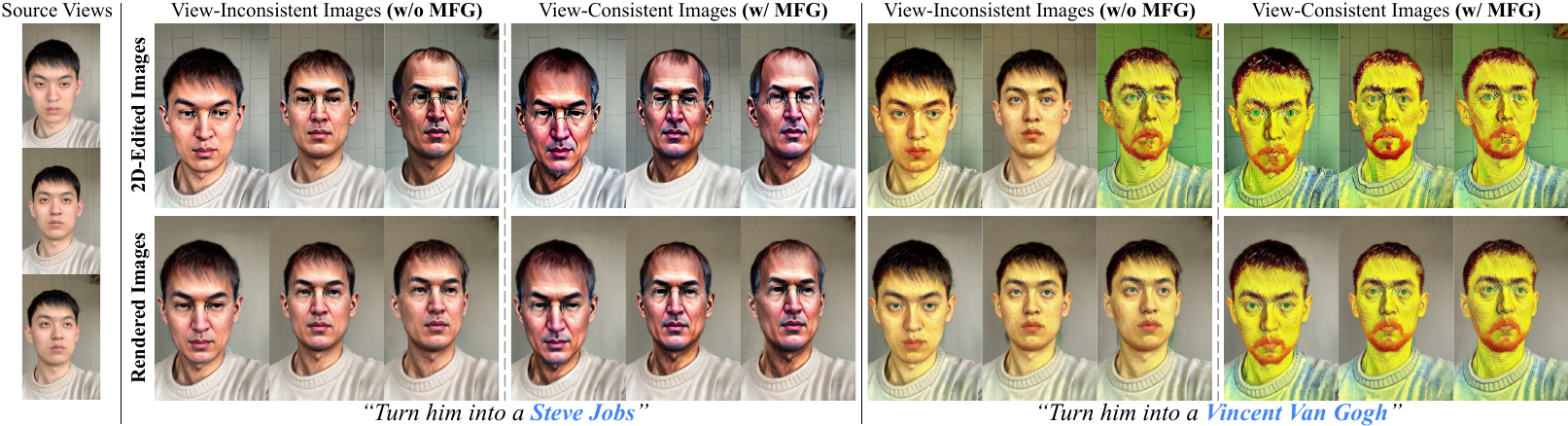}
    \vspace{-2.em}
    \caption{
    \textbf{MFG ablation}. The top row shows the results of 2D editing with and without MFG. With MFG, the 2D diffusion model produces multi-view-consistent results. In the bottom row, we have images rendered from edited 3DGS based on the above images. Editing 3DGS with view-consistent images results in clear outputs with high fidelity, while other cases produce inconsistent results with low fidelity.
    }
    \label{fig:abl_MFG}
    \vspace{-1.8em}
\end{figure*}

\subsection{Implementation Details}
\label{sec:impl_details}

Our EditSplat is built upon vanilla 3DGS~\cite{kerbl20233d} using PyTorch. We use InstructPix2Pix~\cite{brooks2023instructpix2pix} (IP2P) as a 2D editor with our proposed MFG, as described in \ref{sec:MFG}. To validate the 3D scene editing performance of EditSplat, we collect 4 scenes from IN2N~\cite{haque2023instruct}, one scene from BlendedMVS~\cite{yao2020blendedmvs}, and 3 scenes from Mip-NeRF360~\cite{barron2022mip}, covering datasets utilized in previous studies, including complex real-world large 360° scenes. All experiments are conducted on a single RTX A6000 GPU. The complete editing process takes approximately 6 minutes for the ``Face'' scene in IN2N.

\vspace{-1.3em}
\paragraph{Hyperparameters} 
All experiments are conducted with fixed hyperparameters across all scenes and prompts, except for the number of epochs (ranging from 5 to 10). For a fair comparison with baseline methods such as GaussianEditor~\cite{chen2024gaussianeditor} and DGE~\cite{chen2024dge}, which adopt the IP2P~\cite{brooks2023instructpix2pix} and set to \( s_T = 7.5 \), \( s_I = 1.5 \), we set the guidance scales of MFG to \( s_T = 7.5 \), \( s_M = 1.0 \), and \( s_S = 0.5 \). Specifically, we matched the text guidance scale \( s_T \) with the baselines and adjusted the image guidance scales such that the sum of \( s_M \) and \( s_S \) equals \( s_I \), ensuring comparable experimental conditions. In AGT, the threshold value for localization \( w_{thres} \) is set to \( 0.1 \), and the pruning proportion \( k \) is \( 0.15 \).

\subsection{Baselines}
We mainly compare EditSplat with the current state-of-the-art methods that are based on 3DGS like ours: GaussianEditor~\cite{chen2024gaussianeditor}, GaussCtrl~\cite{wu2024gaussctrl}, and DGE~\cite{chen2024dge}. Although both GaussianEditor and DGE employ IP2P as we do, with instruction-based prompts, GaussCtrl utilizes ControlNet~\cite{zhang2023adding}, which uses description-like prompts. To ensure a fair comparison, we generate the source scene descriptions to reproduce GaussCtrl's input conditions. Since these baselines represent the current leading approaches and provide a comprehensive benchmark in 3DGS-based text-driven 3D scene editing, we evaluate the performance of EditSplat in comparison with them. 

\subsection{Qualitative Comparisons}


\cref{fig:teaser} illustrates EditSplat's ability to achieve multi-view consistent editing and elaborate optimization for 3D scene editing. In the first row, EditSplat modifies objects in a 360° scene according to the text prompt, maintaining consistent textures for both ``wild boar" and ``polar bear" across different views. The second row shows EditSplat transforming a man's face into ``Harry Potter" and a ``marble sculpture" with consistent results, despite the geometric complexity of facial features. In the third row, EditSplat demonstrates precise editing of intricate objects, accurately altering leaves by effectively optimizing Gaussians in targeted regions.

In \cref{fig:comparison}, we compare EditSplat with recent editing baselines. Most baselines suffer from suboptimal visual changes or blurring artifacts in rendered images due to multi-view inconsistency of 2D-edited images. In comparison, it is clear that our method gives more pronounced edited results, closely aligning with the given text prompt. 

Notably, in the fourth row, prompted by ``Turn him into a Minecraft character,'' GaussianEditor yields minimal edits with artifacts in the facial region, while DGE and GaussCtrl produce blurry and suboptimal results, respectively. In contrast, our method produces clear outputs, achieving a more authentic Minecraft-like appearance. This highlights that our MFG is more effective guidance for diffusion than depth-based conditioning in GaussCtrl, epipolar constraints in DGE, or iterative dataset updating in GaussianEditor. 

Additionally, EditSplat demonstrates superior performance in global transformations and localized editing for large real-world scenes. Specifically, in the last row with the prompt ``Make it autumn,'' other methods fail to preserve the original table color, whereas our method maintains the table's color while effectively modifying the surrounding foliage. This effectiveness arises from AGT's selective optimization of Gaussians based on the semantics of the attention map corresponding to the prompt token ``autumn,'' enabling precise and context-aware semantic localization.

\subsection{Quantitative Comparisons}
Despite the subjective nature of 3D scene generation and editing, we adhere to established practices~\cite{wang2025view, wu2024gaussctrl, chen2024gaussianeditor, park2023ed, he2024customize, koo2024posterior, chen2024dge} by employing CLIP~\cite{radford2021learning} for quantitative evaluation. Our comparison metrics include two types of CLIP scores. CLIP text-image directional similarity captures the alignment between changes in text captions and the corresponding changes between rendered and source images in CLIP space. Meanwhile, CLIP text-image similarity measures the alignment between the CLIP embeddings of the target text and the rendered images. In addition, we perform user study based on human preference. The user study was conducted with 100 participants via Amazon Mechanical Turk~\cite{amazonturk}. All of our metrics are evaluated under 8 scenes with 2 prompts each to ensure fair comparison and diverse feedback. As shown in Table \ref{tab:Quantitative}, EditSplat achieves the highest scores across all categories, demonstrating superior performance. This demonstrates that EditSplat yields results that most closely align with the text prompt while achieving the highest human perceptual quality. Details are provided in the supplementary materials.



\subsection{Ablation Study}

\begin{table}[t]
\footnotesize
\vspace{-1.65em}
\centering
\tabcolsep=0.2cm
\resizebox{\columnwidth}{!}
{
\begin{tabular}{ l c c c c c c }
    \toprule
    Method & CLIP\textsubscript{dir} & CLIP\textsubscript{sim} & User Study \\
    \specialrule{0em}{0pt}{1pt}
    \hline
    \specialrule{0em}{0pt}{2pt}
    GaussianEditor~\cite{chen2024gaussianeditor} & 0.0923 & 0.2243 & 0.1643 \\
    GaussCtrl~\cite{wu2024gaussctrl} & 0.0979 & 0.2336 & 0.1987 \\
    DGE~\cite{chen2024dge} & 0.1222 & 0.2394 & 0.2143 \\
    \specialrule{0em}{0pt}{1pt}
    \hline
    \specialrule{0em}{0pt}{2pt}
    EditSplat (Ours) & \textbf{0.1431} & \textbf{0.2531} & \textbf{0.4227} \\
    \bottomrule
\end{tabular}}
\vspace{-1.em}
\caption{
\textbf{Quantitative Comparison}. 
CLIP\textsubscript{dir}: CLIP text-image direction similarity; CLIP\textsubscript{sim}: CLIP text-image similarity. User study conducted to evaluate human preference.
}
\vspace{-2.em}
\label{tab:Quantitative}
\end{table}

To demonstrate the effectiveness of our proposed components: Multi-view Fusion Guidance (MFG) and Attention-Guided Trimming (AGT), we conducted an ablation study.

\vspace{-1.2em}
\paragraph{Multi-View Fusion Guidance (MFG).}

\cref{fig:intro}-(a) illustrates the impact of MFG on rendered results from edited 3DGS. MFG integrates multi-view information from edited outputs and provides this guidance to the diffusion process, facilitating consistent 2D edits that contribute to reliable 3D editing. Without MFG, inconsistent 2D edits lead to suboptimal outcomes, such as blurring, artifacts, and minimal visual changes in the rendered 3DGS images.

For example, as shown in \cref{fig:abl_MFG}, when prompted with ``Turn him into a Steve Jobs'' and ``Turn him into a Vincent Van Gogh,'' editing without MFG results in minimal visual changes or blurriness and artifacts. Conversely, using MFG produces consistent final renderings that accurately reflect the intended 2D edits, underscoring the importance of MFG in achieving high-quality 3D scene editing.

\vspace{-1.2em}
\paragraph{Attention-Guided Trimming (AGT).}



In \cref{fig:intro}-(b), we demonstrate AGT's effectiveness in handling significant geometric transformations, such as reshaping the man's nose into a circular red clown nose. Applying AGT results in a deeper red nose, showcasing its capability for substantial visual and geometric modifications. 

\begin{figure}[t]
    \vspace{-1.5em}
    \centering
    \includegraphics[width=\linewidth]{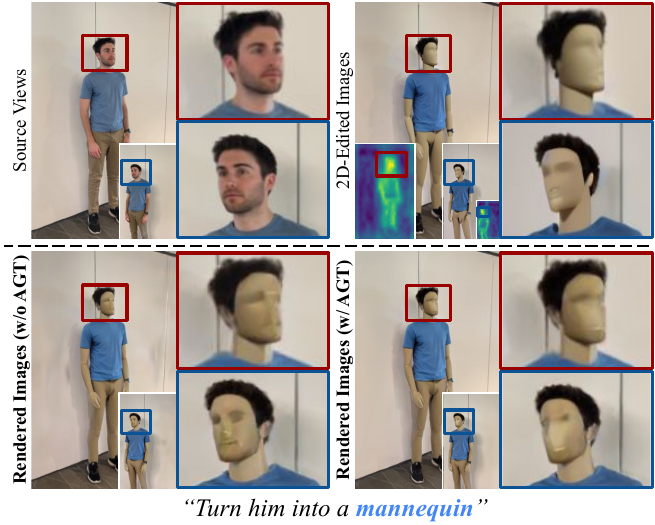}
    \vspace{-2.em}
    \caption{
    \textbf{AGT ablation}. With our proposed AGT, rendered images from edited 3DGS have more significant geometric changes, while others fail. As a result, the outcome aligns more closely with the desired 2D-edited images.
    }
    \label{fig:abl_AGT}
    \vspace{-1.8em}
\end{figure}

At the top-right of 2D edited images in \cref{fig:abl_AGT}, the cross-attention map highlights the head region (in yellow) with high attention weights, indicating areas that require considerable visual and geometric adjustments. Without AGT, pre-trained Gaussians retaining excessive source information persist in these high-attention regions, hindering convergence and leading to suboptimal 3D edits with remnants of the original face shape and artifacts, particularly around the nose. In contrast, AGT prunes these redundant Gaussians, aligning the 3D scene more closely with the editing guidance. This results in rendered images where the face undergoes precise geometric changes that align with the 2D edits, such as transforming into a ``mannequin.''



Moreover, without AGT, local editing lacks control, leading to unintended editing in non-target areas (e.g., the clown's jacket in \cref{fig:intro}-(b) and the background in \cref{fig:abl_AGT}). With AGT, Gaussians with high attention weights are selectively optimized, while those with low attention are excluded, ensuring precise local edits.




\section{Conclusion}
\label{sec:conclusion}

We introduce EditSplat, a robust 3D scene editing framework comprising Multi-view Fusion Guidance (MFG) and Attention-Guided Trimming (AGT). MFG addresses the critical challenge of achieving multi-view consistency, which often results in suboptimal outputs such as minimal edits or blurred results. By integrating critical multi-view information into the diffusion process, MFG ensures consistent and precise edits across different viewpoints, facilitating editing of source images and direct optimization of the 3D Gaussian Splatting model, unlike iterative editing approaches. AGT enhances optimization efficiency and enables precise and semantically rich local editing by pruning pre-trained Gaussians and selectively optimizing those relevant. Extensive evaluations of various complex real-world scenes demonstrate that EditSplat achieves state-of-the-art performance existing methods, setting a new benchmark for text-driven 3D scene editing.



\vspace{0.5em}
\noindent \textbf{Limitations.}
EditSplat relies on the quality of 2D diffusion models~\cite{brooks2023instructpix2pix} and the depth maps rendered by 3DGS~\cite{kerbl20233d}. Current 2D diffusion models may struggle with complex or nuanced prompts, potentially limiting the effectiveness of 3D editing guidance. Additionally, the rendering and depth map quality of 3DGS are still developing, which can affect the projection process in MFG. We expect that advancements in both 2D diffusion models and 3DGS will enhance the robustness and precision of our framework.


\clearpage
\section*{Acknowledgements} {
This research was supported by Korea Research Institute for defense Technology planning and advancement - Grant funded by Defense Acquisition Program Administration(DAPA)((KRIT-CT-23-021), 90\%), 
the Ministry of Science and ICT (MSIT) of Korea, under the National Research Foundation (NRF) grant (RS-2024-00337548, 5\%; RS-2025-00521602, 1\%),
the Culture, Sports and Tourism R\&D Program through the Korea Creative Content Agency grant funded by the Ministry of Culture, Sports and Tourism in 2024~(International Collaborative Research and Global Talent Development for the Development of Copyright Management and Protection Technologies for Generative AI, RS-2024-00345025, 3\%),
Institute of Information \& communications Technology Planning \& Evaluation (IITP) grant funded by the Korea government(MSIT) (No. RS-2019-II190079, Artificial Intelligence Graduate School Program(Korea University), 1\%), and 
Artificial intelligence industrial convergence cluster development project funded by the Ministry of Science and ICT(MSIT, Korea)\&Gwangju Metropolitan City.}

{
    \small
    \bibliographystyle{ieeenat_fullname}
    \bibliography{main}
}

\maketitlesupplementary
\setcounter{page}{1}
\setcounter{section}{0}
\renewcommand{\thesection}{\Alph{section}}

\newcommand{\myparagraph}[1]{\vspace{2pt}\noindent{\bf #1}}
\newcommand\framework{{EditSplat}}

\section*{Overview}

This supplementary material introduces further details and experimental results of our proposed method, EditSplat.

\begin{itemize}

\item \cref{sec:supp_result} introduces additional results of EditSplat, including comparison of CLIP score trends over iteration with baselines.
\vspace{0.5em}

\item \cref{sec:supp_mfg} provides a detailed explanation of Multi-View Fusion Guidance (MFG), including the multi-view fusion process and the formulation of MFG.
\vspace{0.5em}

\item \cref{sec:supp_agt} elaborates on Attention-Guided Trimming (AGT) details, including preparing attention maps and further analysis of ablation study on pruning method, a key component of AGT, across iterations.
\vspace{0.5em}

\item \cref{sec:supp_ex} outlines the experimental setup, including implementation details and user study.

\end{itemize}

\vspace{-1.em}
\section{Additional Results}
\label{sec:supp_result}
\subsection{Extensive Results}
\begin{figure*}[t]
    \vspace{-1.5em}
    \centering
    \includegraphics[width=\linewidth]{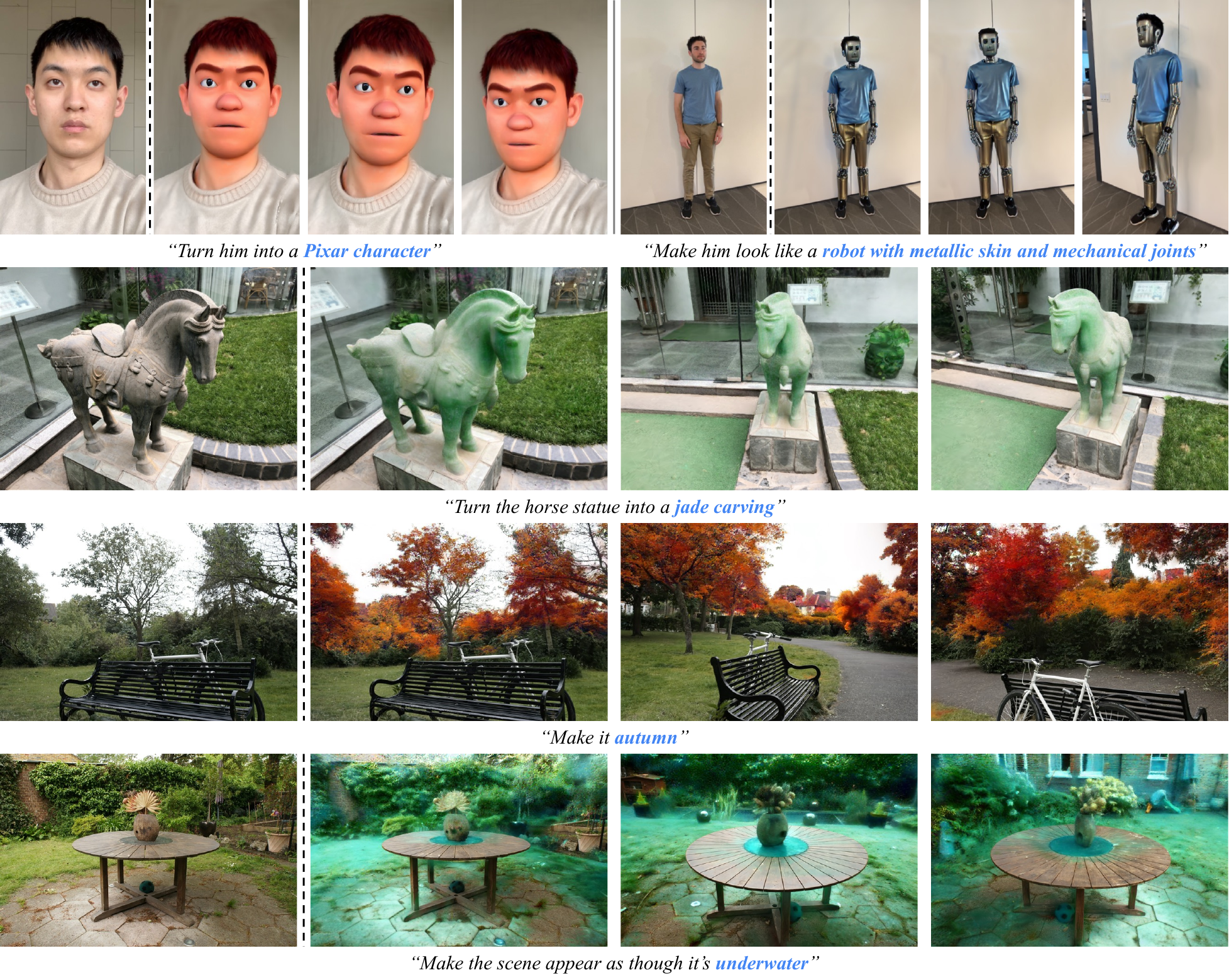}
    \vspace{-2.em}
    \caption{
    \textbf{Additional extensive results 1}.
    We present extensive qualitative results to highlight the robustness and versatility of our proposed method. Our EditSplat ensures multi-view consistency and provides flexible editing, ranging from fine-grained modifications to global stylization.
    }
    \label{fig:extensive_1}
    \vspace{-1.8em}
\end{figure*}

\begin{figure*}[t]
    \vspace{-1.5em}
    \centering
    \includegraphics[width=\linewidth]{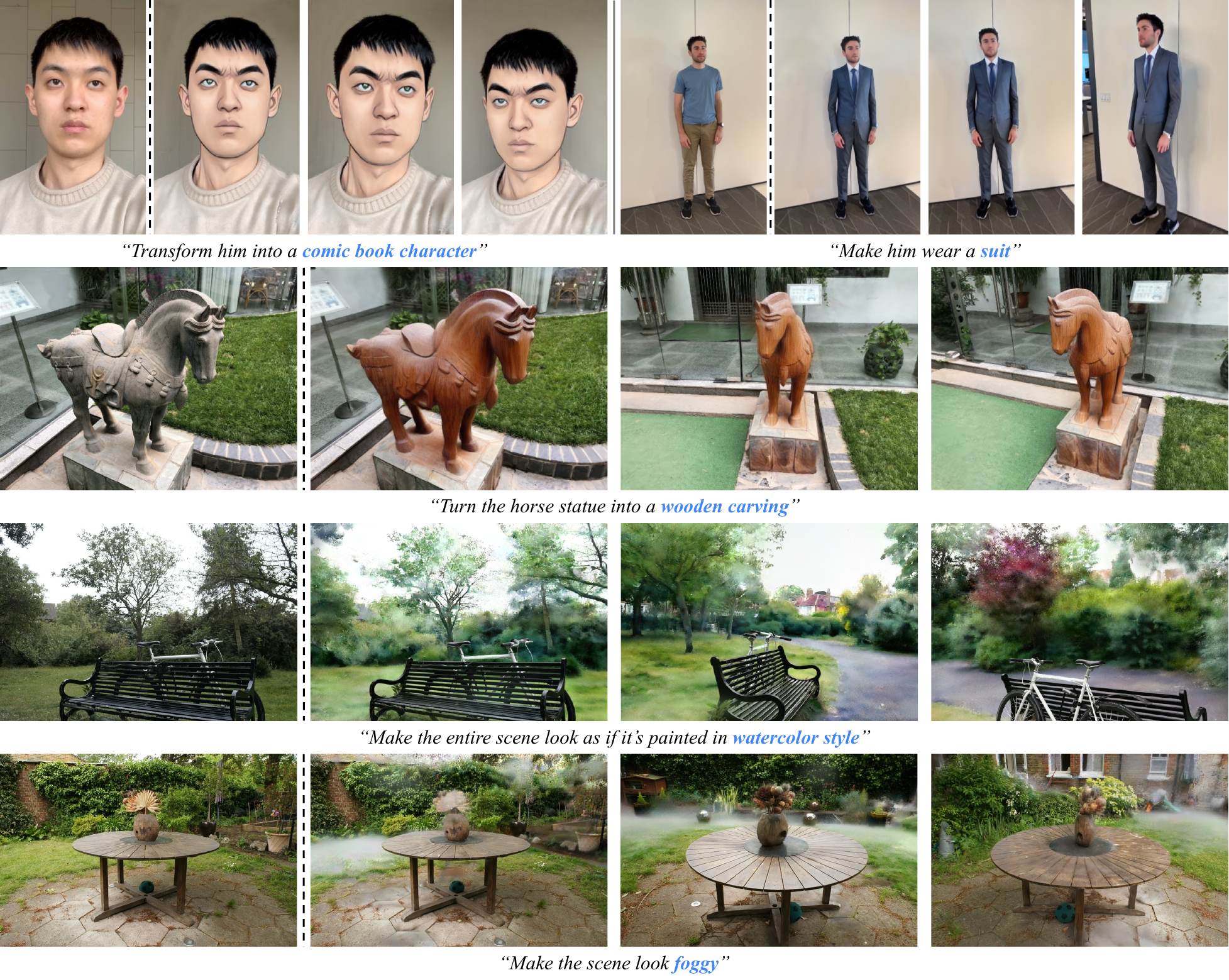}
    \vspace{-2.em}
    \caption{
    \textbf{Additional extensive results 2}.
    We present extensive qualitative results to highlight the robustness and versatility of our proposed method. Our EditSplat ensures multi-view consistency and provides flexible editing, ranging from fine-grained modifications to global stylization.
    }
    \label{fig:extensive_2}
    \vspace{-1.8em}
\end{figure*}

We present extensive results to demonstrate the capability of EditSplat to handle a variety of scenarios, including large-scale scenes and complex text instructions, as shown in \cref{fig:extensive_1} and \cref{fig:extensive_2}. 

\myparagraph{Video Results and Supplementary Files.}
To further demonstrate our method with additional results not included in the main and this supplementary paper, we provide rendered videos and a project page. 

\begin{figure}[t]
    \vspace{-0.3em}
    \centering
    \includegraphics[width=\linewidth]{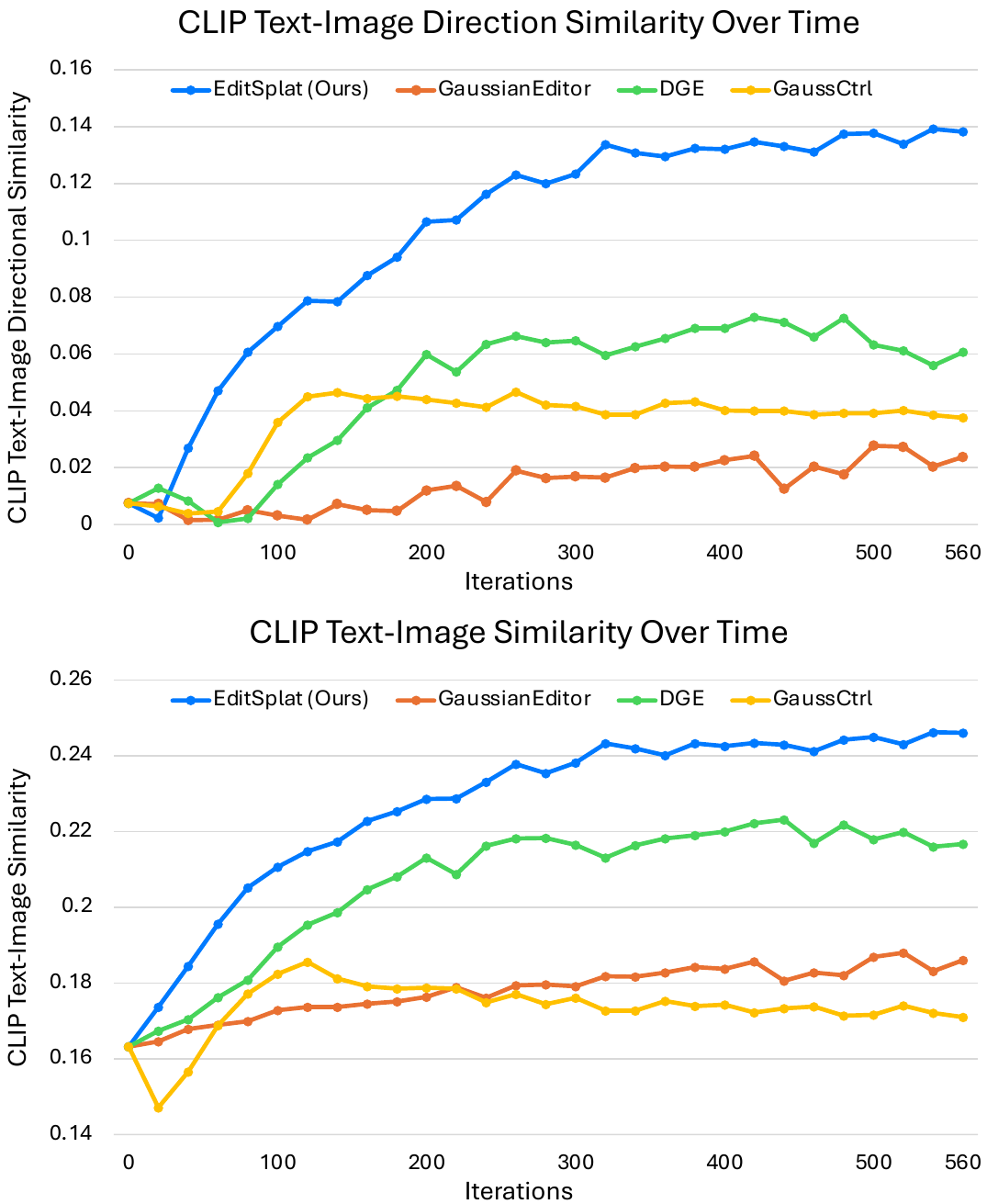}
    \vspace{-1.8em}
    \caption{
    \textbf{Comparison of CLIP Scores Across Iterations.}.
    This figure compares the CLIP similarities between EditSplat and other baseline models over iterations. The results highlight the superior effectiveness of EditSplat in maintaining semantic alignment with the given instructions.
    }
    \label{fig:clip_graph}
    \vspace{-1.5em}
\end{figure}

\myparagraph{Comparison of CLIP Score Trends.}
We present a graph in \cref{fig:clip_graph} that illustrates the trends of CLIP~\cite{radford2021learning} text-image directional similarity and CLIP text-image similarity across iterations for baselines on the ``Face" scene in IN2N~\cite{haque2023instruct} with text prompt ``Make his face resemble that of a marble sculpture." The graph highlights the optimization effectiveness and performance trajectory of EditSplat compared to the baselines throughout iterations.

The results demonstrate that EditSplat achieves superior semantic alignment with the given instructions and improved optimization efficiency compared to the baselines. Both EditSplat's CLIP text-image directional similarity and CLIP text-image similarity scores are the highest and increase significantly faster, indicating superior convergence efficiency. These results suggest that the AGT technique improves optimization efficiency, while the MFG editing process ensures multi-view consistency.

\begin{figure*}[!t]
    \vspace{-1.5em}
    \centering
    \includegraphics[width=\linewidth]{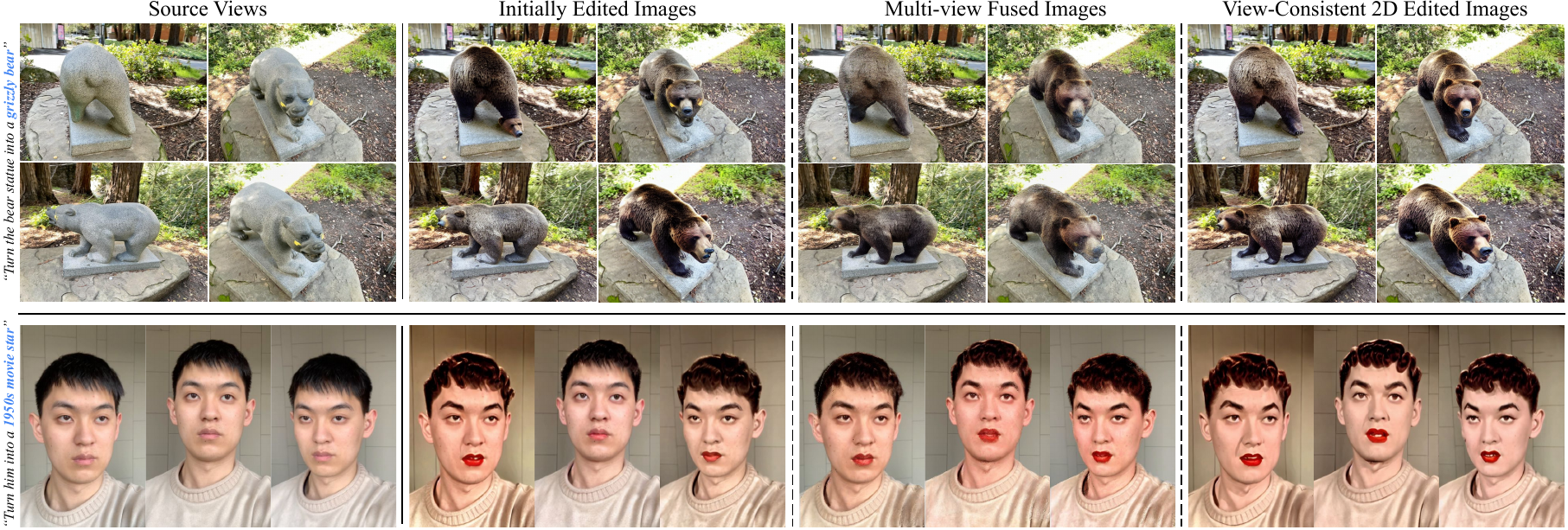}
    \vspace{-2.em}
    \caption{
    \textbf{Multi-View Fusion Guidance (MFG) 2D Editing Process.} MFG resolves multi-view inconsistency in the initially edited images through a multi-view fusion process that integrates information from multiple views to produce multi-view fused images. These fused images are subsequently incorporated into the diffusion process, ensuring consistent editing across all views.
    }
    \label{fig:supp_MFG}
    \vspace{-1.8em}
\end{figure*}

\subsection{Comparison with NeRF-based Method}

\begin{figure*}[t]
    \vspace{-1.em}
    \centering
    \includegraphics[width=\linewidth]{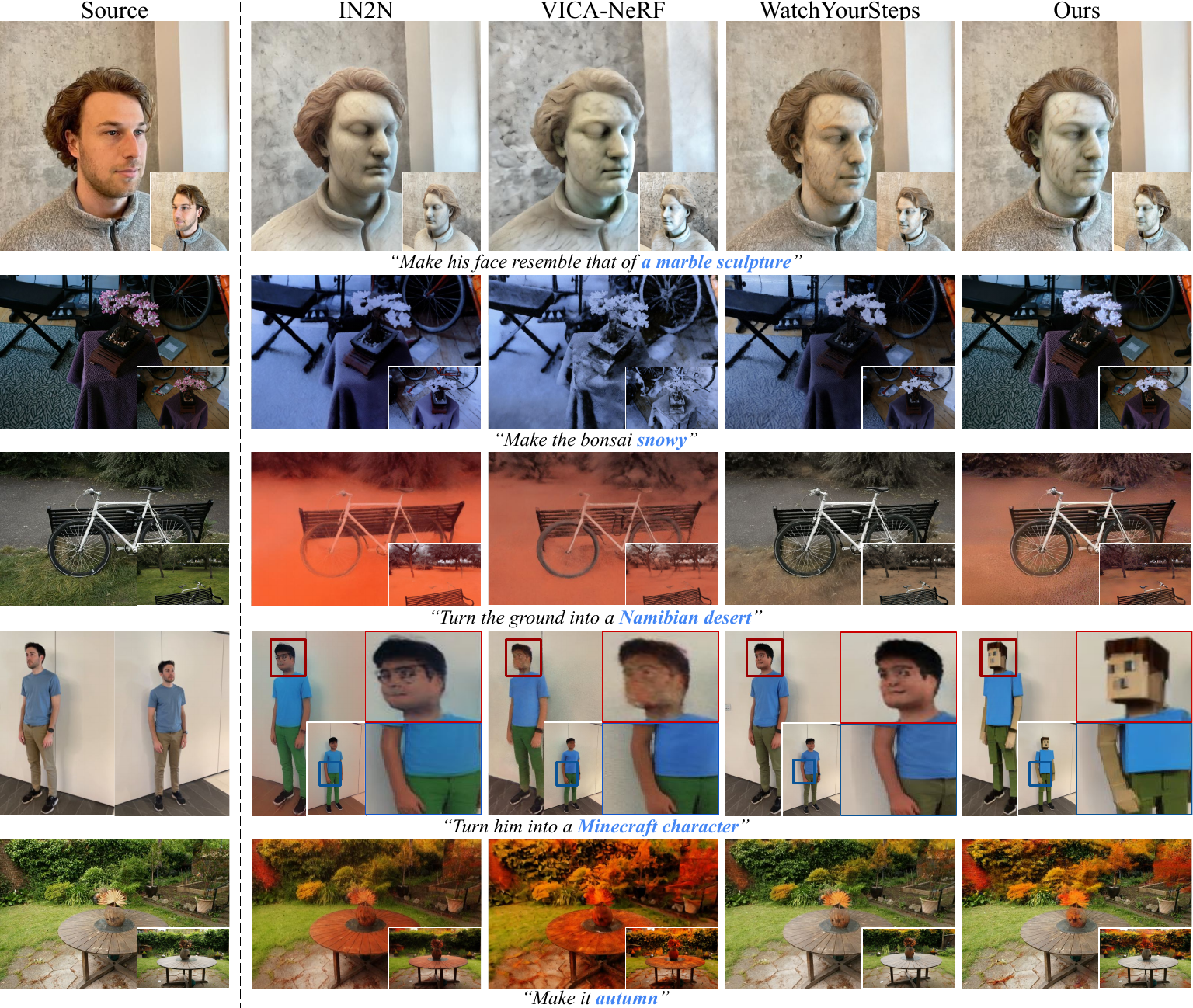}
    \vspace{-2.em}
    \caption{
    \textbf{Qualitative Comparison with NeRF-Based Methods.} EditSplat provides more intense and precise editing compared to recent NeRF-based methods. The leftmost column displays the source images, while the subsequent columns show rendered images. To evaluate multi-view consistency, different views of the corresponding images are included in each corner. Notably, EditSplat demonstrates superior performance in both local and global editing tasks.
    }
    \label{fig:supp_comparison}
    \vspace{-1.5em}
\end{figure*}

We compare our EditSplat framework with recent Neural Radiance Fields (NeRF)\cite{mildenhall2021nerf}-based approaches that were not included in the main paper's baselines, as those focused on state-of-the-art models that utilize 3D Gaussian Splatting (3DGS)\cite{kerbl3Dgaussians}. Specifically, we evaluate EditSplat against InstructNeRF2NeRF (IN2N)\cite{haque2023instruct}, Vica-NeRF\cite{dong2024vica}, and WatchYourSteps~\cite{mirzaei2025watch}, as illustrated in \cref{fig:supp_comparison}.


The results demonstrate that NeRF-based methods produce outputs that are less aligned with target prompts and exhibit inferior editing quality compared to EditSplat. These methods often suffer from blurriness, artifacts, and minimal edits, with limited capability for precise local editing. In contrast, EditSplat achieves clear and high-quality rendered results with accurate local and global edits. Moreover, EditSplat completes the 3D editing process for the ``Face" scene in the IN2N dataset in approximately 6 minutes, while NeRF-based methods require over 50 minutes, with WatchYourSteps taking nearly four hours. These comparisons underscore EditSplat's superior efficiency, higher-quality edits, and better alignment with target prompts.

\vspace{-0.5em}
\section{Multi-View Fusion Guidance (MFG) Details}
\label{sec:supp_mfg}


\subsection{Multi-View Fusion}
\label{sec:supp_fusion}


\begin{figure*}[!t]
    \vspace{-1.5em}
    \centering
    \includegraphics[width=\linewidth]{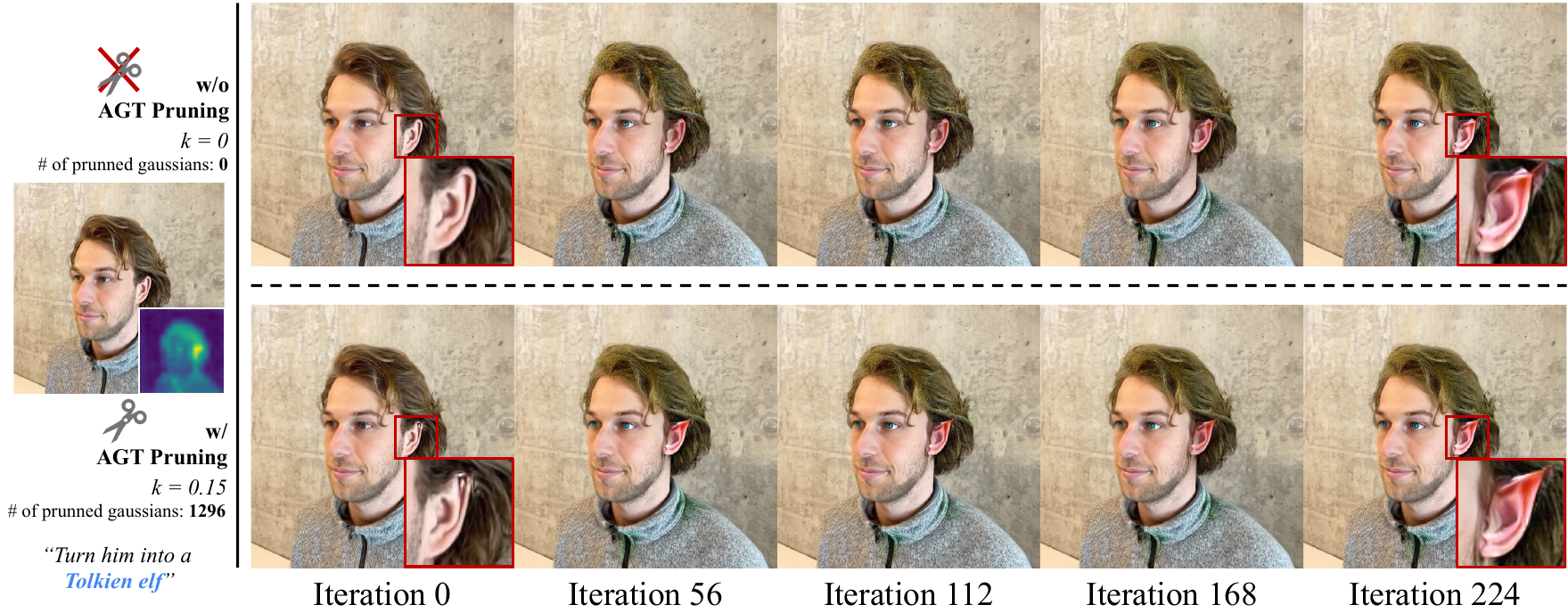}
    \vspace{-2.em}
    \caption{
    \textbf{Ablation Study on Pruning in AGT}.
    The top row illustrates the inefficiency of 3DGS optimization during editing. High-attention regions, which require significant modifications, remain under-edited due to the presence of numerous source Gaussians. In contrast, the application of pruning through AGT effectively removes Gaussians in high-attention areas, enabling more accurate and efficient editing. As demonstrated in the bottom row, this approach allows for better optimization even with fewer iterations.
    }
    \label{fig:pruning}
    \vspace{-1.8em}
\end{figure*}

\begin{figure}[t]
    \vspace{-1.em}
    \centering
    \includegraphics[width=0.86\linewidth]{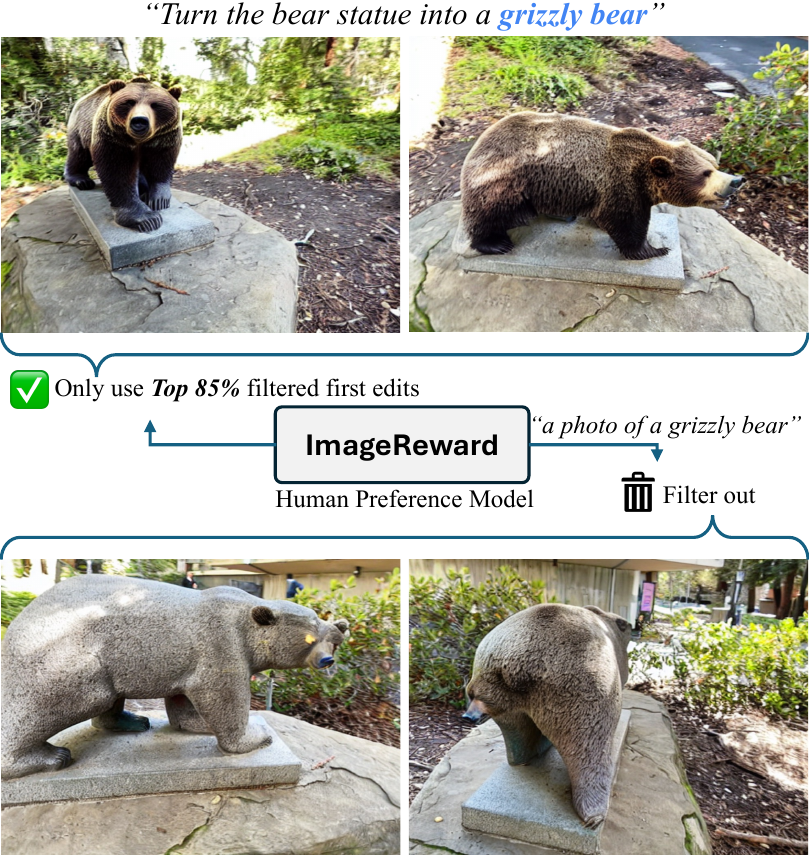}
    \vspace{-0.8em}
    \caption{
    \textbf{Filtering} 
    To prevent misaligned images from negatively affecting MFG, we leverage ImageReward~\cite{xu2024imagereward} to rank the initially edited images. Only the top 85\% of these images, based on their rankings, are utilized for MFG, ensuring high-quality inputs.
    }
    \label{fig:reward}
    \vspace{-1.5em}
\end{figure}

\vspace{-0.5em}
\myparagraph{Filtering.}
The ImageReward model~\cite{xu2024imagereward} is trained on 137k pairs of expert comparisons and evaluates image fidelity and text-image alignment based on human preferences. This model's scoring capabilities surpass those of CLIP~\cite{radford2021learning}, Aesthetic~\cite{schuhmann2022laion}, and BLIP~\cite{li2022blip}, as demonstrated in the original paper. We leverage the ImageReward model to rank the initially edited images by their alignment with the text prompt and select the top 85\% of high-scoring images. This filtering step improves the quality of the multi-view fused images, ensuring more accurate integration across views. We provide the ImageReward's capability in our filtering process in \cref{fig:reward}.

\myparagraph{Depth-based Multi-View Fusion.}
Following the filtering step, we project the top-ranked initially edited images to each target view. Specifically, we select the top 5 adjacent views based on proximity to the target view. A naive projection from multiple views to a single target view leads to suboptimal results, including inaccuracies and blurriness caused by the improper handling of overlapping pixels. To address this, we employ an iterative alpha blending strategy guided by depth values, enhancing both consistency and accuracy in the fused results. The algorithm for multi-view fusion is detailed in \cref{alg:fusion}. In \cref{alg:fusion}, the input selected source data \(\{(\mathcal{I}_{\text{src}}^i, \mathcal{D}_{\text{src}}^i, \xi_{\text{src}}^i, K_{\text{src}}^i)\}_{i=1}^{N'}\) represents the selected initially edited images, their corresponding depth maps from 3DGS~\cite{kerbl20233d}, extrinsic camera parameters, and intrinsic camera parameters, respectively. The target camera parameters \(\{(\xi_{\text{trg}}^j, K_{\text{trg}}^j)\}_{j=1}^{N}\) define the extrinsic \(\xi_{\text{trg}}^j\) and intrinsic \(K_{\text{trg}}^j\) parameters for each of the \(N\) target views. Here, \(N\) represents the total number of views, while \(N'\) denotes the number of selected images.
\vspace{1.em}

\begin{algorithm}[!ht] 

\SetAlgoLined
\SetKwInOut{Input}{Input}
\SetKwInOut{Output}{Output}

\Input{
    Selected source data \(\{(\mathcal{I}_{\text{src}}^i, \mathcal{D}_{\text{src}}^i, \xi_{\text{src}}^i, K_{\text{src}}^i)\}_{i=1}^{N'}\),
    Target camera parameters \( \{(\xi_{\text{trg}}^j, K_{\text{trg}}^j)\}_{j=1}^{N}\)
}

\Output{
    Multi-view fused images \(\{ \mathcal{I}_{\text{trg}}^j \in \mathbb{R}^{H \times W \times 3} \}_{j=1}^{N}\)
}

\BlankLine
\ForEach{target view \(j = 1\) to \(N\)}{
    \textbf{Select Nearest Source Views:}
    
    \(\mathcal{I}_{\text{sel}} \leftarrow\) Select 5 nearest views to \(\xi_{\text{trg}}^j\) from \(\{ \mathcal{I}_{\text{src}}^i \}_{i=1}^{N'}\)

    \BlankLine
    \textbf{Initialization:}
    
    \(\mathcal{I}_{\text{trg}}^j \leftarrow \mathbf{0}\) tensor of size \((3, H, W)\)

    \BlankLine
    \textbf{Reprojection:}
    
    \ForEach{source view \(i\) in \(\mathcal{I}_{\text{sel}}\)}{
        \tcp{\(P_i\): 3D points, \(C_i\): RGB colors.}
        \((P_i, C_i) \in \mathbb{R}^{HW \times 3}, u_i \in \mathbb{R}^{HW \times 2}\)\\
        \((P_i, C_i)  \mkern-1mu \leftarrow \text{Reproject}(\mathcal{I}_{\text{src}}^i, \mathcal{D}_{\text{src}}^i, \xi_{\text{src}}^i, K_{\text{src}}^i, \xi_{\text{trg}}^j, K_{\text{trg}}^j)\)
        
        \(u_i   \leftarrow \text{MapToPixelCoordinates}(P_i)\)
        
        \tcp{\(u_i\) within image bounds.}
        \tcp{\(\mathcal{I}_{\text{list}}, \mathcal{D}_{\text{list}}\): List of RGB, depth.}
        
        \(\mathcal{I}_{\text{list}}[i][u_i] \leftarrow C_i,\)
        \(\mathcal{D}_{\text{list}}[i][u_i] \leftarrow \text{depth of } P_i\)
    }

    \BlankLine
    \textbf{Blending Based on Depth:}
    
    Sort \((\mathcal{D}_{\text{list}}, \mathcal{I}_{\text{list}})\) in descending order by \(\mathcal{D}_{\text{list}}\).

    \ForEach{\((\mathcal{D}_l, \mathcal{I}_l)\) in \((\mathcal{D}_{\text{list}}, \mathcal{I}_{\text{list}})\)}{
        \eIf{\(l = 0\)}{
            \(w \leftarrow 1\)
        }{
            \(w \leftarrow \frac{\mathcal{D}_l}{\mathcal{D}_l + \mathcal{D}_{prev}}\)
        }
        \(\mathcal{I}_{\text{trg}}^j \leftarrow (1-w) \cdot \mathcal{I}_l + w \cdot \mathcal{I}_{\text{trg}}^j\)
    }

}
\Return{\(\{ \mathcal{I}_{\text{trg}}^j \}_{j=1}^{N}\)}
\caption{Depth-based Multi-View Fusion in MFG}
\label{alg:fusion}
\end{algorithm}

\myparagraph{Background Refinement.}
The initial multi-view fused images \(I_{trg}\) often exhibit a sparse background, while the target editing object in the image, which is often located in the center of the scene, appears dense. This issue arises due to the limitations of reprojection caused by discrepancies in camera viewpoints. To address this, we refine \(I_{trg}\) by replacing its sparse background with the background from the corresponding source image, using SAM~\cite{kirillov2023segment} to preserve the original background. First, as the source image's object is generally misaligned with the target object's range, we extract a binary mask \(M_{trg}\) of the target object from \(I_{trg}\):
\vspace{-0.1em}
\begin{equation}
    M_{trg}(x, y) =
\begin{cases} 
1 & \text{if } (x, y) \text{ belongs to the target object,} \\
0 & \text{otherwise}.
\end{cases}
\end{equation}
Using \(M_{trg}\), we isolate the background from the source image  \(\mathcal{I}_{\text{src}}\) as:
\vspace{-0.5em}
\begin{equation}
    B_{src}(x, y) = \mathcal{I}_{\text{src}}(x, y) \cdot (1 - M_T(x, y)).
\end{equation}
Finally, the refined multi-view fused image \(h_M\) is obtained by combining the target object from \(\mathcal{I}_{\text{trg}}\) with the background from \(\mathcal{I}_{\text{src}}\), ensuring seamless integration:
\vspace{-0.5em}
\begin{equation}
    h_M(x, y) =
\begin{cases}
\mathcal{I}_{\text{trg}}(x, y), & \text{if } M_{trg}(x, y) = 1, \\
\mathcal{I}_{\text{src}}(x, y), & \text{if } M_{trg}(x, y) = 0.
\end{cases}
\end{equation}
This process effectively replaces the sparse background in \(\mathcal{I}_{\text{trg}}\) while preserving the target object, resulting in smoother and more cohesive multi-view fused images \(h_M\). 

We illustrate the intermediate results of the MFG editing process in \cref{fig:supp_MFG}. The initially edited images exhibit misaligned edits with the text prompt and lack consistency across views. In contrast, the multi-view fused images produced through the multi-view fusion process are well-aligned with the text prompt, consistent across views, and incorporate multi-view information. Finally, the 2D edited images ensure multi-view consistency and precise alignment with the text prompt.


\subsection{Alignment with Multi-View Information}
\label{sec:supp_guidance}

To seamlessly incorporate multi-view information during the editing process, we extend the classifier-free guidance method. This allows the integration of multi-view fusion details into the diffusion process to facilitate multi-view consistent editing. Below is our Multi-view Fusion Guidance, an extended score estimate for multi-view aligned editing with classifier-free guidance, as specified in the main paper:

\begin{small}
\begin{equation}
\begin{aligned}
    \tilde{\epsilon_\theta}\left(z_t, h_S, h_T, h_M\right) & =  \; {\epsilon_\theta}\left(z_t, \varnothing, \varnothing\right) \\
    & + s_T   \left({\epsilon_\theta}\left(z_t, h_M, h_T\right) - {\epsilon_\theta}\left(z_t, h_M, \varnothing\right)\right) \\
    & + s_M   \left({\epsilon_\theta}\left(z_t, h_M, \varnothing\right) - {\epsilon_\theta}\left(z_t, \varnothing, \varnothing\right)\right) \\
    & + s_S  \left({\epsilon_\theta}\left(z_t, h_S, \varnothing\right) - {\epsilon_\theta}\left(z_t, \varnothing, \varnothing\right)\right),
\end{aligned}
\label{eq:supp_MFG_same}
\end{equation}
\end{small}
Here, \( h_M \) represents the multi-view fusion image, \( h_S \) is the source image, and \( h_T \) corresponds to the text prompt. The guidance strength for each conditioning is modulated by the respective scale factors \( s_M \), \( s_S \), and \( s_T \).

\myparagraph{Conditional Probability Formulation.}
In our approach, the conditional probability distribution \( P(z | h_M, h_S, h_T) \) can be expressed as:
\vspace{-0.5em}
\begin{align}
P(&z | h_M, h_S, h_T) =  \frac{P(z, h_M, h_S, h_T)}{P(h_M, h_S, h_T)} \nonumber \\[0.2cm]
= & \; \frac{P(h_T | h_M, h_S, z) P(h_M | h_S, z) P(h_S | z) P(z)}{P(h_M, h_S, h_T)}
\end{align}
\vspace{1.em}

\myparagraph{Log Probability and Score Estimation.}
Taking the logarithm of the conditional probability results in:

\vspace{-1.8em}
\begin{align}
\log(P(z | h_M, h_S, h_T)) & = \log(P(h_T | h_M, h_S, z)) \nonumber \\
& + \log(P(h_M | h_S, z)) \nonumber \\
& + \log(P(h_S | z)) \nonumber \\
& + \log(P(z)) \nonumber \\
& - \log(P(h_M, h_S, h_T))
\end{align}
\textbf{}Calculating the gradient with respect to \( z \) and rearranging terms, we obtain:

\vspace{-1.8em}
\begin{align}
\nabla_z \log(P(z | h_M, h_S, h_T)) & = \nabla_z \log(P(z)) \nonumber \\
& + \nabla_z \log(P(h_S | z)) \nonumber \\
& + \nabla_z \log(P(h_M | h_S, z)) \nonumber \\
& + \nabla_z \log(P(h_T | h_M, h_S, z)) 
\end{align}
\vspace{-1.8em}

\begin{figure}[t]
    \vspace{-0.5em}
    \centering
    \includegraphics[width=0.9\linewidth]{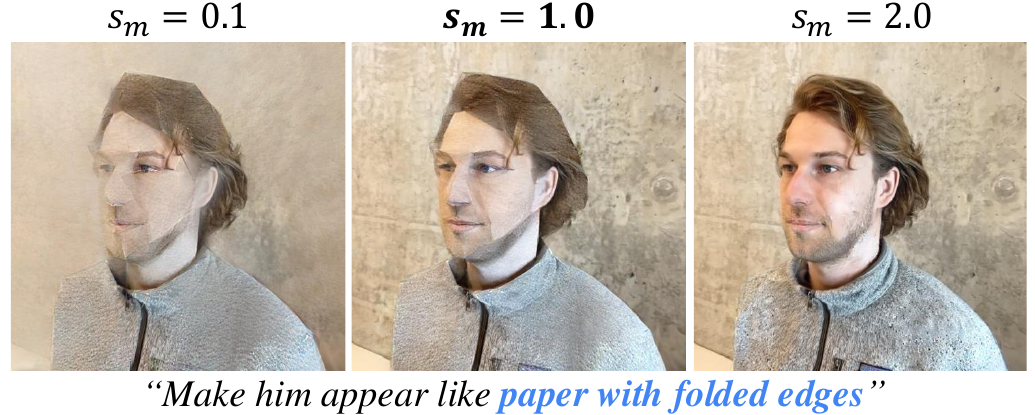}
    \vspace{-0.8em}
    \caption{
    \textbf{Influence of Guidance Scale \( s_M \).} Qualitative analysis of \( s_M \) in MFG for 3D editing. Higher \( s_M \) improves multi-view consistency, but excessively large value reduces the editing effect.
    }
    \label{fig:MFG_s_M}
    \vspace{-1.5em}
\end{figure}

\myparagraph{Guidance Interpretation.}
Each guidance scale (e.g., \( s_M \), \( s_S \), and \( s_T \)) effectively shifts the probability mass toward outputs that align with the corresponding conditioning. For instance, \( s_M \) biases the implicit classifier \( p_{\theta} \) toward assigning higher probabilities to multi-view information, thereby ensuring consistent 2D editing across views and enhancing the quality of the resulting 3D edits (see \cref{fig:MFG_s_M}). However, excessively increasing \( s_M \) reduces the influence of the text guidance scale \( s_T \), resulting in less pronounced editing effects. Conversely, \( s_S \) preserves the original content from the source image, and \( s_T \) promotes adherence to the provided text prompts. By carefully balancing these guidance scales, our model effectively achieves multi-view consistent edits that accurately reflect both structural and color details specified by textual instructions.

\begin{figure}[t]
    \centering
    \includegraphics[width=0.9\linewidth]{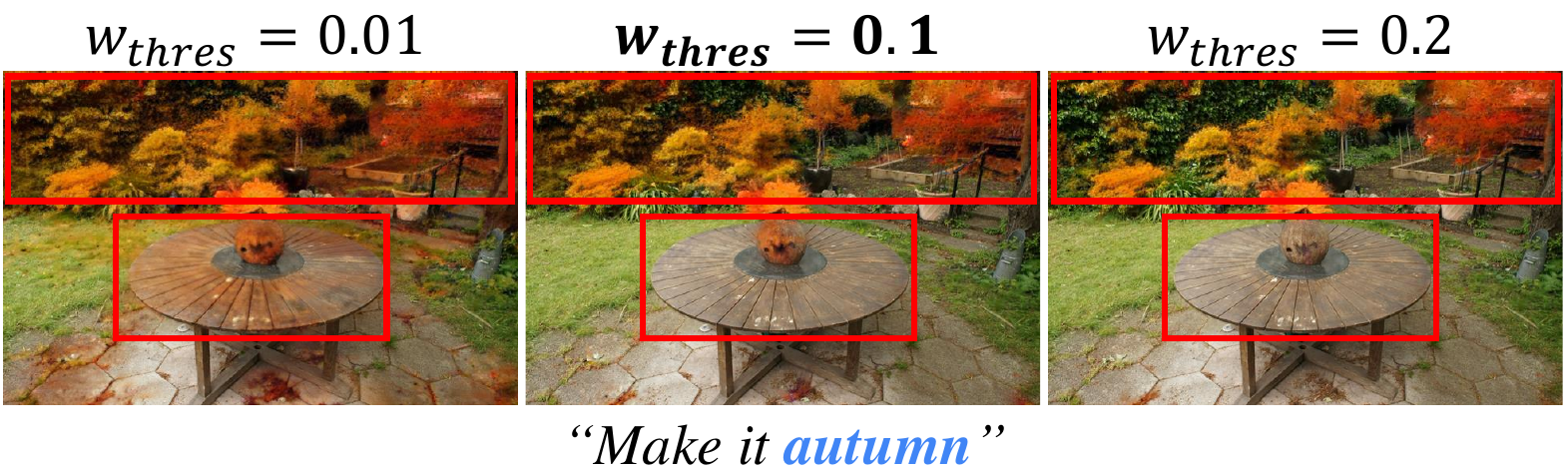}
    \vspace{-0.8em}
    \caption{
    \textbf{Influence of \(w_{thres}\)}. Qualitative comparison of \( w_{thres} \) in AGT for local editing. Lower \(w_{thres}\) fails to preserve original structures, while higher values overly restrict the editing scope.
    }
    \label{fig:AGT_w_thres}
    \vspace{-1.5em}
\end{figure}

\section{Attention-Guided Trimming (AGT) Details}
\label{sec:supp_agt}


\begin{figure}[t]
    \vspace{-1.em}
    \centering
    \includegraphics[width=0.9\linewidth]{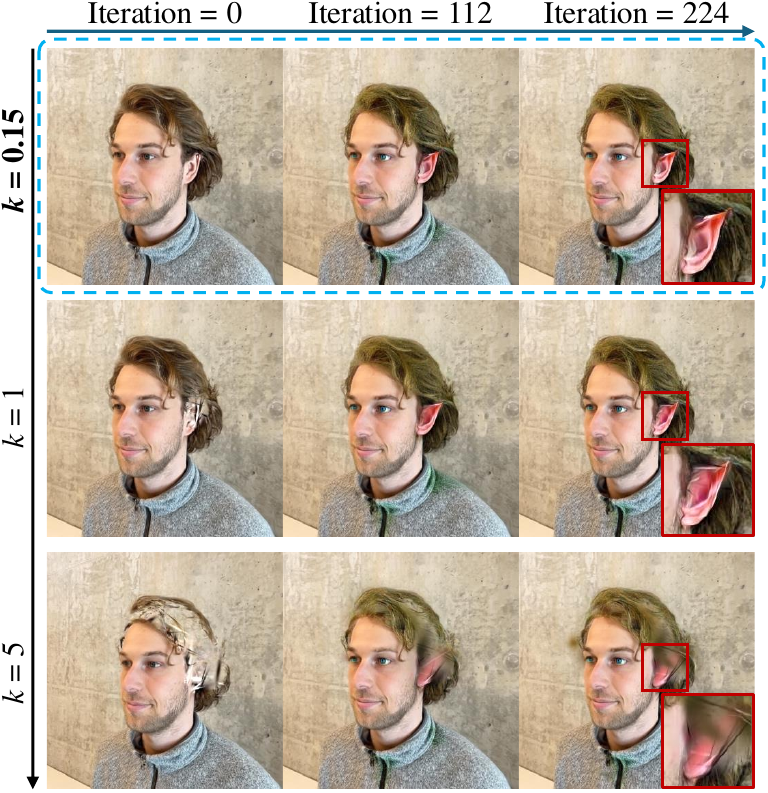}
    \vspace{-0.8em}
    \caption{
    \textbf{Ablation study on pruning threshold \textit{k} in AGT}.
    The figure illustrates the training progress under different pruning thresholds \textit{k} using the same source images and text prompt as in \cref{fig:pruning}. Excessively high values of \textit{k} result in the removal of too many Gaussians, hindering the editing process.
    }
    \label{fig:prun_iter}
    \vspace{-1.5em}
\end{figure}

\subsection{Extracting Attention Map}
When the latent representation of the input image added noise and the text prompt are fed into the diffusion model for denoising, we select the semantic keyword that represents the intended editing outcome (e.g., ``autumn" in the instruction ``Make it autumn" or ``clown" in ``Turn him into a clown"). We then extract all cross-attention maps associated with this keyword, which are computed during the MFG editing process. These attention maps are resized to rendering resolution using bilinear interpolation, aggregated, and normalized to the [0, 1] range using Min-Max normalization. This process allows us to accurately assign these semantic maps to each Gaussian, ensuring they contain meaningful regions for pruning, facilitating efficient optimization, and selectively optimizing semantically rich local editing.

\subsection{Qualitative Analysis over Iterations}
We further validate the effectiveness of AGT by analyzing the editing results across iterations. Notably, the ablation study presented here analyzes the effect of pruning independently within AGT while maintaining its selective optimization for local editing.

\cref{fig:pruning} highlights the contrast in edited results with and without pruning the Gaussians. Based on the instruction \textit{``Turn him into a Tolkien elf"}, the attention map on the far left identifies the ear as the most prominent region, requiring significant modifications in the source scene. For optimal editing, our AGT first assigns attention weights to each source Gaussians. Then, the top 0.15\% of Gaussians (1,296 in total) based on assigned weights, are pruned. The rendered image in the bottom row at iteration 0 shows that the upper part of the ear has been cleared. As a result, the ear is edited more effectively into the desired elf ear. In contrast, the top row shows that the remaining source Gaussians in the unpruned case interfere with the convergence. 

\cref{fig:prun_iter} illustrates how editing quality varies with different pruning threshold \textit{k}\%. Especially in the third row, pruning the top 5\% of Gaussians results in excessively empty regions, requiring additional iterations to fill these gaps. Conversely, when \textit{k}\% is appropriately set to 0.15\%, optimal editing is achieved. Through heuristic analysis across various scenes, we find that pruning the top 0.15\% of Gaussians is the most suitable approach for overall scenes. 

In addition, both \cref{fig:pruning} and \cref{fig:prun_iter} demonstrate that the attention weights assigned by AGT effectively reflect the semantic importance of the editing regions. Given the text instruction \textit{``Turn him into a Tolkien elf"}, the ear, rather than the head, should undergo more significant changes. Our results highlight that our AGT accurately reflects this semantic importance during editing and that pruning an appropriate percentage of Gaussians ensures effective convergence toward the target.

\subsection{Analysis of Local Editing}

We provide an analysis of local editing in Attention-Guided Trimming (AGT), focusing specifically on the impact of the threshold \( w_{thres} \) for semantic local editing. As illustrated in \cref{fig:AGT_w_thres}, varying \( w_{thres} \) significantly affects the balance between preserving original scene details and effectively applying the intended semantic edits.

When the threshold \( w_{thres} \) is set too low, attention-guided selection becomes excessively inclusive, causing unintended modifications beyond the target area. Consequently, critical structural elements of the original scene—such as the table in \cref{fig:AGT_w_thres}—fail to be properly preserved.
In contrast, higher values of \( w_{thres} \) significantly narrowing the scope of editing. Although this helps maintain structural integrity by protecting essential regions, it simultaneously limits the effectiveness of the semantic editing. As demonstrated in our qualitative evaluation, the edits within regions of secondary semantic importance (e.g., grass or background foliage in \cref{fig:AGT_w_thres}) become overly constrained, resulting in less pronounced visual changes and diminishing the overall impact of the edit.

Therefore, selecting an appropriate \( w_{thres} \) is critical to achieving an optimal trade-off between preserving essential structures of the original scene and effectively implementing localized semantic editing. Our experiments reveal that intermediate values of \( w_{thres} \) successfully balance these competing objectives, enabling precise and semantic localized editing.






\section{Experiment Set Up}
\label{sec:supp_ex}

\subsection{Implementation Details}

EditSplat utilizes the vanilla 3DGS~\cite{kerbl20233d} framework for 3D reconstruction. For each scene, we train the model for 30,000 iterations to serve as the source scene. All 3DGS training phases employ the Adam optimizer~\cite{kingma2014adam}, an identical learning rate with vanilla 3DGS. We apply the same densification strategy across all scenes, with a densification interval of 100 and a gradient threshold of 0.01.

For the 2D image editor, we use InstructPix2Pix(IP2P)~\cite{brooks2023instructpix2pix} from Diffusers library~\cite{von-platen-etal-2022-diffusers} with our novel method, MFG. In a more detailed setup, we perform 20 sampling steps using the DDIM scheduler~\cite{song2021denoising}, with noise sampled from $t \in [0.7, \ 0.98]$. 



\subsection{Evaluation}

\myparagraph{Quantitative.} 
We adopt a train/test split for our datasets following the methodology suggested by Mip-NeRF360~\cite{barron2022mip}, taking every 8th image for test. Our evaluation metrics include measuring text-image directional similarity and text-image similarity using CLIP~\cite{radford2021learning}. The text descriptions used for calculating CLIP similarity are detailed in \cref{tab:clip}. In \cref{tab:prompts}, we summarize the instructions employed for 3D editing. For one of our comparison baselines, GaussCtrl~\cite{wu2024gaussctrl}, which does not support instruction-based editing, \cref{tab:prompts} provides the source and target scene descriptions used in its evaluation. Additionally, since GaussCtrl leverages a different 2D image editor, such as ControlNet~\cite{zhang2023adding}, we set its guidance scale to the recommended value of 5.

\begin{figure*}[t]
    \vspace{-1.5em}
    \centering
    \includegraphics[width=\linewidth]{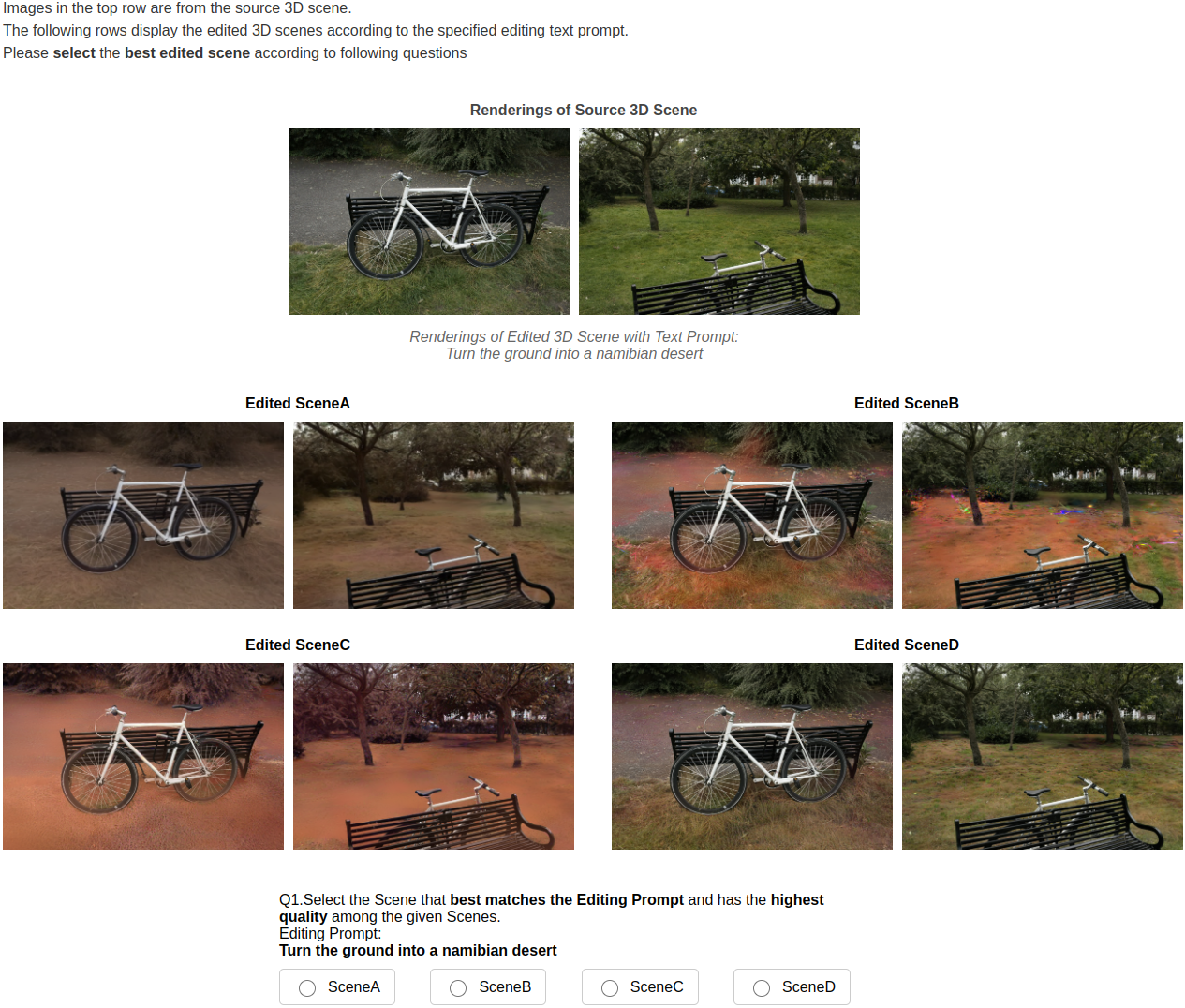}
    \vspace{-2.em}
    \caption{
    \textbf{User Study Survey Form}. Participants are presented with the source 3D scene and four edited scenes, which include results from three baselines and our EditSplat. These edited scenes are randomly shuffled for each question to prevent bias. Participants evaluate the edits based on how well they align with the given text prompt and their overall quality.
    }
    \label{fig:user_study}
    \vspace{-1.8em}
\end{figure*}

\myparagraph{User Study.}
To ensure a fair and rigorous user study, we recruit 100 participants through Amazon Mechanical Turk~\cite{amazonturk}, a widely-used platform for human evaluation. Participants are provided with images rendered from the source scene along with the corresponding text prompt used for editing. They are tasked with evaluating how well the edited images align with the given text prompt and assessing the overall quality of the edits. The choices include rendered views edited by our method and those generated by baseline models, all presented in a randomized order to prevent participants from inferring which model produced which images. Furthermore, the randomization is applied separately for each scene to ensure unbiased evaluation across different examples. This setup allows us to measure both the semantic accuracy of the edits and the perceived quality of the rendered images, as shown in \cref{fig:user_study}.


\begin{table*}[ht] 
\renewcommand{\arraystretch}{1.2} 
\centering 
\begin{tabular}{ >{\centering\arraybackslash}p{2cm}  
                >{\centering\arraybackslash}p{3cm}  
                >{\centering\arraybackslash}p{3cm}  
                >{\centering\arraybackslash}p{5.8cm} } 
    \hline
    \textbf{Scene} & \textbf{Source Description} & \textbf{Target Description} & \textbf{Editing Instruction} \\ \hline
    
    bear & a photo of a bear & a photo of a wild boar & Turn the bear statue into a wild boar\\ 
    bear & a photo of a bear & a photo of a metallic robot bear & Turn the bear statue into a metallic robot\\ \hline
    bicycle & a photo of a park & a photo of a Namibian desert & Turn the ground into a Namibian desert\\ 
    bicycle & a photo of a park & a watercolor style paint of a park & Make the entire scene look as if it's painted in watercolor style\\ \hline
    bonsai & a photo of a plant & a photo of a plant, snowy & Make the bonsai snowy\\ 
    bonsai & a photo of a plant & a photo of a plant made of paper & Change the bonsai to look like it's made of paper, folded into intricate origami shapes\\ \hline
    face & a photo of a face of a man & a photo of a marble sculpture & Make his face resemble that of a marble sculpture\\ 
    face & a photo of a face of a man & a photo of a face of a man, made of paper  & Make him appear like he’s made of paper with folded edges\\ \hline
    fangzhou & a photo of a face of a man & a photo of a face of Jocker & Turn him into a Jocker\\ 
    fangzhou & a photo of a face of a man & a photo of a face of Steve Jobs & Turn him into a Steve Jobs\\ \hline
    garden & a photo of an outdoor garden & a photo of an outdoor garden in autumn & Make it autumn\\ 
    garden & a photo of an outdoor garden & a photo of a garden in underwater & Make the scene appear as though it's underwater\\ \hline
    person & a photo of a person & a photo of a person wearing a suit & Make him wear a suit\\ 
    person & a photo of a person & a photo of a person in Minecraft & Turn him into a Minecraft character\\ \hline
    stone horse & a photo of a horse statue & a photo of a horse made of wood & Turn the horse statue into a wooden carving\\ 
    stone horse & a photo of a horse statue & a photo of a horse made of jade & Turn the stone horse into a jade carving\\ 
    \hline
\end{tabular}
\caption{
    \textbf{Details of the descriptions used for the CLIP similarity}. CLIP similarity is computed as the cosine similarity between embeddings in the CLIP space. The source description depicts the scene before editing, while the target description represents the desired edited scene. Both descriptions are transformed into text embeddings in the CLIP space and are used to evaluate the semantic alignment of the 3D scene.
}
\label{tab:clip}
\end{table*}

\begin{table*}[ht] 
\renewcommand{\arraystretch}{1.2} 
\centering 
\begin{tabular}{ >{\centering\arraybackslash}p{2cm}  
                >{\centering\arraybackslash}p{5.8cm}  
                >{\centering\arraybackslash}p{3cm}  
                >{\centering\arraybackslash}p{3cm} } 
    \hline
    \textbf{Scene} & \textbf{Editing Instruction} & \textbf{Source Description} & \textbf{Target Description} \\ \hline
    bear & Turn the bear statue into a wild boar & a photo of a bear statue in the forest & a photo of a wild boar in the forest \\ 
    bear & Turn the bear statue into a metallic robot & a photo of a bear statue in the forest & a photo of a metallic robot in the forest\\ \hline
    bicycle & Turn the ground into a Namibian desert & a photo of a bicycle at grass & a photo of the bicycle at the namibian desert \\ 
    bicycle & Make the entire scene look as if it's painted in watercolor style & a photo of a bicycle at grass & a photo of a bicycle scene as if it's painted in watercolor style \\ \hline
    bonsai & Make the bonsai snowy & a photo of a bonsai in the desk & a photo of a snowy bonsai \\ 
    bonsai & Change the bonsai to look like it's made of paper, folded into intricate origami shapes & a photo of a bonsai in the desk & a photo of a tree made of paer, folded into intricate origianl shapes \\ \hline
    face & Make his face resemble that of a marble sculpture & a photo of a face of a man & a photo of a marble sculpture \\ 
    face & Make him appear like he’s made of paper with folded edges & a photo of a face of a man & a photo of a man made of paper with folded edges \\ \hline
    fangzhou & Turn him into a Jocker & a photo of a face of a man & a photo of a Joker \\ 
    fangzhou & Turn him into a Steve Jobs & a photo of a face of a man & a photo of a Steve Jobs \\ \hline
    garden & Make it autumn & a photo of a fake plant on a table in the garden & a photo of a garden scene with autumn \\ 
    garden & Make the scene appear as though it's underwater & a photo of a fake plant on a table in the garden & a photo of a garden scene appear as though underwater \\ \hline
    person & Make him wear a suit & a photo of a person & a photo of a man wearing a suit \\ 
    person & Turn him into a Minecraft character & a photo of a person & a photo of a Minecraft character \\ \hline
    stone horse & Turn the horse statue into a wooden carving & a photo of a stone horse statue in front of the museum & a photo of a wooden carving statue in front of the museum \\ 
    stone horse & Turn the stone horse into a jade carving & a photo of a stone horse statue in front of the museum & a photo of a stone horse of a jade carving\\ 
    \hline
\end{tabular}
\caption{
\textbf{Detailed text prompts used for 3D scene editing}. While models using IP2P as a 2D image editor can perform editing directly based on instructions, model employing inversion-based 2D image editor such as GaussCtrl~\cite{wu2024gaussctrl} requires both source and target descriptions. To ensure a fair comparison, we designed the source and target descriptions to ensure that the semantic difference between source and target accurately reflects the given editing instructions.
}
\label{tab:prompts}
\end{table*}

\end{document}